\documentclass[journal]{IEEEtran}
\usepackage{times}
\usepackage{epsfig}
\usepackage{graphicx}
\usepackage{amsmath}
\usepackage{amssymb}
\usepackage{epstopdf}
\DeclareUnicodeCharacter{FF0C}{fi}

\begin{document}

\title{Multi-view 3D Reconstruction with Transformer}

\author{Dan~Wang, Xinrui~Cui$\dagger$, Xun~Chen, Zhengxia~Zou, Tianyang~Shi,\\
Septimiu Salcudean \IEEEmembership{Fellow,~IEEE}, Z.~Jane~Wang \IEEEmembership{Fellow,~IEEE}, and Rabab Ward \IEEEmembership{Fellow,~IEEE}
\thanks{$\dagger$ Corresponding author: Xinrui Cui (xinruic@ece.ubc.ca).}

\thanks{D. Wang, X. Cui, Septimiu Salcudean, Z.~Jane~Wang, and Rabab Ward are with University of British Columbia, Canada. e-mail: \{danw, xinruic, Tims, zjanew, rababw\}@ece.ubc.ca.
X. Chen is with University of Science and Technology of China.
e-mail: xunchen@ustc.edu.cn.
Z. Zou is with University of Michigan, Ann Arbor.
e-mail: zzhengxi@umich.edu.
T. Shi is with NetEase Fuxi AI Lab. e-mail: shitianyang@corp.netease.com.
}}

\maketitle

\begin{abstract}
Deep CNN-based methods have so far achieved the state of the art results in multi-view 3D object reconstruction. Despite the considerable progress, the two core modules of these methods - multi-view feature extraction and fusion, are usually investigated separately, and the object relations in different views are rarely explored. In this paper, inspired by the recent great success in self-attention-based Transformer models, we reformulate the multi-view 3D reconstruction as a sequence-to-sequence prediction problem and propose a new framework named 3D Volume Transformer (VolT) for such a task. Unlike previous CNN-based methods using a separate design, we unify the feature extraction and view fusion in a single Transformer network. A natural advantage of our design lies in the exploration of view-to-view relationships using self-attention among multiple unordered inputs. On ShapeNet - a large-scale 3D reconstruction benchmark dataset, our method achieves a new state-of-the-art accuracy in multi-view reconstruction with fewer parameters ($70\%$ less) than other CNN-based methods. Experimental results also suggest the strong scaling capability of our method. Our code will be made publicly available.




\end{abstract}

\section{Introduction}
\label{sec:intro}

Learning 3D object representation from multi-view images is a fundamental and challenging problem in 3D modeling, virtual reality, and computer animation. Recently, deep learning approaches have greatly promoted the research in multi-view 3D reconstruction problem, where the deep convolutional neural network (CNN) based approaches have so far achieved state-of-the-art accuracy in this task \cite{xie2019pix2vox, yang2020robust, xie2020pix2vox++}. 

To learn effective 3D representation from multiple input views, most recent CNN-based approaches follow the design principle of divide-and-conquer, where a common practice is to introduce a CNN for feature extraction and fusion module for integrating the features or reconstruction results from multiple views. Despite the strong connection between the two modules, their methodology designs are investigated separately. Also, during the CNN feature extraction stage, the object relations in different views are rarely explored. Although some recent approaches have introduced recurrent neural network (RNN) to learn object relationships between different views~\cite{choy20163d,NIPS2017_9c838d2e}, such a design lacks computational efficiency and the input views to the RNN model are permutation-sensitive \cite{44871}, which is difficult to be compatible with a set of unordered input views. It is also shown in recent researches that CNN-based reconstruction methods may suffer from the model scaling problem. For example, when the number of model inputs exceeds a certain scale (e.g. 4), the accuracy of the model will be saturated, showing the difficulty of learning complementary knowledge through a large set of independent CNN feature extraction units \cite{yang2020robust, xie2020pix2vox++}.


\begin{figure*}
	\centering
	\includegraphics[width=1\linewidth]{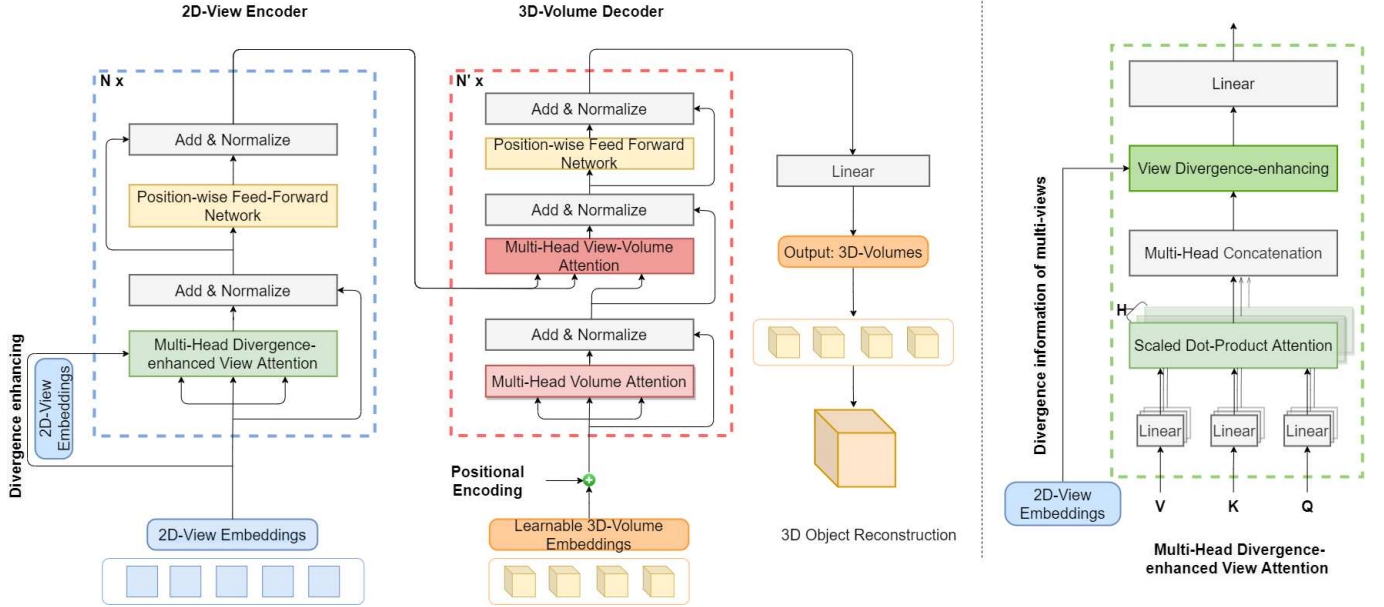}
	\centering
\caption{Illustration of EVolT for Multi-view 3D Object Reconstruction (left).
The proposed view-divergence enhancing function in our EVolT (right).}
	\label{fig:flow}
\end{figure*}


Considering the above challenges, we propose a new framework named ``3D Volume Transformer (VolT)'', and explore the potential of recent great success of self-attention based language model for the multi-view 3D object reconstruction task. We reformulate the multi-view 3D reconstruction as a sequence-to-sequence prediction problem and unify the feature extraction and view fusion in a single Transformer network. On one hand, in multi-view 3D reconstruction, from a particular view, we can only see part of the underlying 3D structure. On the other hand, in a Transformer model, the self-attention mechanism has recently shown its great power on learning complex semantic abstractions within an arbitrary number of input tokens~\cite{devlin2018bert, touvron2020training} and is naturally suitable for exploring the view-to-view relationships of a 3D object's different semantic parts. Given all this, the structure of Transformer \cite{NIPS2017_3f5ee243, dosovitskiy2020image} becomes a natural and attractive solution for the multi-view 3D reconstruction. 

Our Transformer-based framework contains a 2D-view Transformer encoder and a 3D-volume Transformer decoder, as presented in Figure \ref{fig:flow}. In the proposed framework, the 2D-view encoder encodes and fuses the multiple 2D-view information by exploring their ``2D-view $\times$ 2D-view'' relationships of the different inputs. The 3D-volume decoder decodes and fuses the multi-view features from the encoder and generate a 3D probabilistic voxel output for each of the spatial query tokens. The attention layers in the decoder learns ``2D-view $\times$ 3D-volume'' relationships between each of the output voxel grids and input views. Meanwhile, volume attention layers in the decoder further learn ``3D-volume $\times$ 3D-volume'' relationships by exploiting correlations amongst different 3D locations. By using the above unified design, the ``2D-view $\times$ 2D-view'', ``3D-volume $\times$ 3D-volume'', and ``2D-view $\times$ 3D-volume'' relationships can be jointly explored by multiple attention layers in both the encoder and decoder networks.

On basis of the above encoder-decoder structure design, we further investigate the ``attention uniformity'' problem in a Transformer model and propose a effective solution for enhancing the effectiveness of a Transformer model in the multi-view reconstruction task. In Transformers, self-attention possesses a solid inductive bias towards ``token uniformity'' \cite{dong2021attention}, which encourages feature representations of input tokens converge. However, this convergence may further cause the problem of ``attention uniformity'' in deeper layers, which makes a Transformer model loses expressive power speedily with respect to network depth \cite{dong2021attention}. We show that in the multi-view 3D reconstruction task, this problem is particularly prominent and will limit the Transformer's capability of exploring and abstracting multi-view associations at a deeper level. In our experiment, we found that when directly adopting the vanilla Transformer~\cite{NIPS2017_3f5ee243} as our backbone for multi-view 3D reconstruction, the increase in depth will cause degradation of reconstruction accuracy when the model exceeds a certain depth. To tackle it, we further propose the divergence-enhanced Transformer that can slow down the divergence decay in the self-attention layers by enhancing the discrepancy of the embeddings from different views.


The contributions can be summarized as follows:
\begin{itemize}
\item We propose a brand new framework VolT for multi-view 3D object reconstruction. Different from the previous CNN based methods that using a separate design of feature extraction + view fusion, we unify these two stages into a single Transformer network and re-frame the 3D reconstruction as a ``sequence-to-sequence'' prediction problem.
\item  The proposed method can jointly and naturally explore multi-level correspondence and associations between the 2D input views and 3D output volume with in a single unified framework. 
\item We investigate the problem of ``divergence decay'' in the proposed 3D Volume Transformer layers and propose a view-divergence enhancing operation in our self-attention layers to avoid such degradation.
\item Our method achieves a new state-of-the-art for multi-view 3D reconstruction on ShapeNet with only $30\%$ amount of parameters of other recent CNN-based methods. Our method also shows better scaling capability on the number of input views.
\end{itemize}


\section{Related Work} \label{relate}

\subsection{Multi-view 3D Reconstruction}

Reconstructing an object's 3D shape from multi-view images has long been a research hot-spot in both computer vision and computer graphics. Traditional methods \cite{Onur2017,fuentes2015visual} of this field are typically designed based on hand-crafted geometric features. Some representatives of early methods like Structure from Motion (SfM)~\cite{Onur2017}, Simultaneous Localization and Mapping (SLAM)~\cite{fuentes2015visual} need extrinsic camera parameters, which are not always feasible to obtain. Recently, CNN-based approaches, without requiring viewpoint labels or camera calibration, have quickly become the main stream of multi-view 3D reconstruction \cite{choy20163d,NIPS2017_9c838d2e} and have achieved the state of the art reconstruction accuracy. 

In CNN-based methods, a 2D-CNN view encoder, a 3D-CNN view decoder, and a multi-view fusion model are usually separately designed for 3D reconstruction. Among them, the fusion plays an central role in the integration of multi-view feature information. Previous multi-view fusion methods can be roughly grouped into three categories, i.e., pooling-based fusion, learnable weighted-sum fusion and RNN-based fusion. The The pooling-based fusion, including max-pooling fusion and average-pooling fusion, only learns partial information of multiple views and ignores the view associations \cite{huang2018deepmvs, paschalidou2018raynet}. The learnable weighted-sum fusion models are introduced to resolve these problems \cite{xie2019pix2vox, yang2020robust, xie2020pix2vox++}. The RNN-based fusion methods like 3D-R2N2 \cite{choy20163d} and LSM \cite{NIPS2017_9c838d2e} can learn effective view-to-view relations but are computational expensive and permutation-variant \cite{44871}. In this paper, different from the above CNN-based methods, we propose a Transformer-based 3D reconstruction method, which unifies the feature extraction and view fusion in a single model and naturally explore the relationship between different input views.

\subsection{Transformer}

In natural language processing, Transformer models have achieves great success in a variety of tasks such as machine translation, text classification, and question answering \cite{brown2020language}. The key to the Transformer is the multi-head self-attention operation, which aggregates features among every token pairs of the embedding sequence. Recently, Transformer has been also successfully adapted to the computer vision field \cite{carion2020end, dosovitskiy2020image, chen2020generative} and shows promising application prospective. DETR \cite{carion2020end} provides a new framework for object detection that combines a 2D CNN with a Transformer, and directly predicts (in parallel) the final object detection as a sequence of language tokens. ViT \cite{dosovitskiy2020image} applies Transformer directly to sequences of image patches for the image classification task without using CNNs features and have achieves comparable and even higher image classification accuracy when pretrained on large-scale dataset. In CNN-based multi-view 3D reconstruction methods, it is still a difficult task to design a fusion model that can explore the deep relationship between views while maintaining the permutation-invariant capability. A natural advantage of the Transformer in multi-view 3D reconstruction is that its token embedding can be abstracted and learned layer by layer in a disorderly manner, which 
can naturally ease the pain points of CNN-based methods.



\section{Methodology}
\label{sec:methodology}
\subsection{Framework}

The proposed 3D volume Transformer model consists of a 2D-view encoder and a 3D-volume decoder.  The 2D-view encoder encodes the relevant information amongst different views via view attention layers. The 3D-volume decoder learns global correlations of different spatial locations in volume attention layers and predicts the final 3D volume output. We uniformly split the 3D space into a set of tokens and the predicted volumes for each token are finally stitched into the final 3D reconstruction output. Fig.~\ref{fig:flow} shows an overview of the proposed framework.

In this paper, we implement three different versions of method based on the proposed framework:  Vanilla 3D Volume Transformer (\textbf{VolT}), Vanilla 3D Volume Transformer+ (\textbf{VolT+}), and view-divergence-Enhance 3D Volume Transformer (\textbf{EVolT}). 



\begin{itemize}
\item \textbf{VolT:} A baseline implementation of the proposed method using vanilla Transformer model as our baseline and using standard VGG16 features as our initial view embeddings.
\item  \textbf{VolT+:} Using 2D-view embeddings obtained from an advanced pretrained CNN compared with VolT. We use it to testify the impact of 2D-view embeddings on our Transformer-based framework for multi-view 3D reconstruction.
\item \textbf{EVolT:} A full implementation of our method adopting the proposed view-divergence enhancing function in the proposed 3D Volume Transformer framework.
\end{itemize}

Here, to obtain 2D-view initial embeddings, we use a pretrained CNN that is shared among multi-views.
Note that we can also build ``EVolT+'' with an advanced CNN for 2D-view embedding learning in VolT+ to further improve the performance of EVolT.
However, we prefer to keep the small parameter size of the EVolT and emphasize the advantage of the Transformer.


\subsection{Divergence-enhanced 2D-view Encoder}\label{encoder}

Suppose $\mathcal{I}= \{\mathbf{I}^{1}, \mathbf{I}^{2},\cdots ,\mathbf{I}^{M}\}$, where $M$ denote the multi-view image set of an object to be reconstructed. For each $\mathbf{I}^{m}$, we first use a pretrained view-shared CNN to obtain a set of initial view embedding $\mathbf{x}^{m}\in\mathbb{R}^{1\times d}$, where $d$ is the feature dimension. Then, the 2D-view encoder takes in the initial view embeddings $\mathbf{X}_0= [\mathbf{x}^{1}; \mathbf{x}^{2}; \cdots; \mathbf{x}^{M}]\in\mathbb{R}^{M\times d}$ and refines the multi-view representations by exploring a global relationship amongst multiple views using a series of self-attention layers. Here, to keep permutation invariant for the view sequence $\mathcal{X}$, the positional encodings of a standard Transformer are removed. We build our divergence-enhanced 2D-view encoder based on DETR \cite{carion2020end}  by stacking $N=6$ basic blocks. Each basic block consists of a multi-head divergence-enhanced view attention layer (denoted as MH-DEAtt, Eq. (\ref{vi})) and a position-wise feed-forward network (FFN, Eq. (\ref{ffn})). The 2D-view encoder is formulated as follows:
\begin{align}
&\mathbf{X}_0= [\mathbf{x}^{1}; \mathbf{x}^{2}; \cdots; \mathbf{x}^{M}]\\
&\mathbf{X}_{l}=\text{MH-DEAtt}(\mathbf{X}_{l-1},\mathbf{X}_0)\label{vi}\\
&\hat{\mathbf{X}}_{l}=\text{Norm}(\mathbf{X}_{l}+\mathbf{X}_{l-1})\\
&\mathbf{X}_{l}=\text{FFN}(\hat{\mathbf{X}}_{l})\label{ffn}\\
&\mathbf{X}_{l}=\text{Norm}(\mathbf{X}_{l}+\hat{\mathbf{X}}_{l})
\end{align}
where ``Norm'' denotes layer normalization and $l$ is the index of a basic block ($l=1,\cdots,L$). The embeddings of the layer $L$ are used as the output of our 2D-view encoder. 

As shown in the right side of Figure \ref{fig:flow}, the scaled dot-product attention (denoted as Attn) aggregates the feature representations amongst multiple views by learning view-to-view relationships. Meantime, we propose a view-divergence enhancing function (DiView) to ease the discrepancy degradation of the multi-view representations in deeper layers. Specifically, DiView introduces skip connections and concatenates the internal view features with the input view embeddings in the feature dimension. The MH-DEAtt layer is defined as follows
\begin{align}
\text{MH-DEAtt}(\mathbf{X},\mathbf{X}_0)&=\text{DiView}(\mathbf{A},\mathbf{X}_0)\mathbf{W}_{view} \nonumber \\
\text{where } \mathbf{A}&=\text{cat}(\mathbf{A}^{1},\cdots,\mathbf{A}^{H})\\
\mathbf{A}^{h}&=\text{Attn}(\mathbf{Q}^h,\mathbf{K}^h,\mathbf{V}^h) \nonumber
\end{align}
where ``cat'' denotes the concatenation operation and $h$ is the number of head in MH-DEAtt layer. $\mathbf{W}_{view}\in\mathbb{R}^{(Hd_k+d)\times d}$ denotes the parameter matrix of the linear function, and $d_k$ is the feature dimension in each head. In the $h$-th head, $M$ queries stacked in $\mathbf{Q}^h\in\mathbb{R}^{M\times d_k}$ are projected from $M$ view embeddings stacked in $\mathbf{X}$ with the parameter matrix $\mathbf{W}^h_Q\in\mathbb{R}^{d\times d_k}$.
Similarly, the keys and values stacked in $\mathbf{K}^h\in\mathbb{R}^{M\times d_{k}}$ and $\mathbf{V}^h\in\mathbb{R}^{M\times d_{k}}$ are obtained with parameter matrices $\mathbf{W}^h_K\in\mathbb{R}^{d\times d_{k}}, \mathbf{W}^h_V\in\mathbb{R}^{d\times d_{k}}$, respectively:
\begin{align}
\mathbf{Q}^h=\mathbf{X}\mathbf{W}^h_Q; \ \ \mathbf{K}^h=\mathbf{X}\mathbf{W}^h_K;  \ \ \mathbf{V}^h=\mathbf{X}\mathbf{W}^h_V.
\end{align}



Specifically, in the Attention function ``Attn'', the output for a query is represented as an attention-score weighted sum of the values $\mathbf{V}$.
Therefore, the Attn function is formulated as
\begin{equation}
\begin{split}
&\text{Attn}(\mathbf{Q}, \mathbf{K}, \mathbf{V}) = \text{softmax}(\frac{\mathbf{Q}\mathbf{K}^T}{\sqrt{d_{k}}}) \mathbf{V},
\end{split}
\end{equation}
where $d_{k}$ is a scalar for normalization.


\subsection{3D-volume Decoder}

The 3D-volume decoder in our framework learns the global correlation amongst different spatial locations and explore the relationship between the view and spatial domains. Given an object, we denote $\mathbf{Y}_0=[\mathbf{y}^{1}; \mathbf{y}^{2}; \cdots; \mathbf{y}^{N}]$ as a sequence learnable 3D-volume queries at the input end of the decoder, where $\mathbf{y}^n\in\mathbb{R}^{1\times d}$ corresponds to the $n$-th 3D-volume. Here, positional encodings $\mathbf{E}^{pos}$ are added to 3D-volume embeddings to keep the position information in the spatial domain. In the decoder, a basic block contains a volume attention layer, a view-volume attention layer, and a FFN. The decoder can be formulated as follows: 
\begin{align}
&\mathbf{Y}_0= [\mathbf{y}^{1}; \mathbf{y}^{2}; \cdots; \mathbf{y}^{N}]+\mathbf{E}^{pos}\\
&\mathbf{Y}_{l}=\text{MH-VolAttn}(\mathbf{Y}_{l-1})\label{sp}\\
&\mathbf{\hat{Y}}_{l}=\text{Norm}(\mathbf{Y}_{l}+\mathbf{Y}_{l-1})\\
&\mathbf{Y}_{l}=\text{MH-ViewVolAttn}(\mathbf{\hat{Y}}_{l}, \mathbf{X}_{L})\label{vs}\\
&\mathbf{\tilde{Y}}_{l}=\text{Norm}(\mathbf{Y}_{l}+\hat{\mathbf{Y}}_{l})\\
&\mathbf{Y}_{l}=\text{FFN}(\mathbf{\tilde{Y}}_{l})\\
&\mathbf{Y}_{l}=\text{Norm}(\mathbf{Y}_{l}+\mathbf{\tilde{Y}}_{l})
\end{align}
where MH-VolAttn (in Eq.(\ref{sp})) and MH-ViewVolAttn (in Eq.(\ref{vs})) denote the multi-head volume attention layer and the multi-head view-volume attention layer, respectively. We use the embeddings at the layer $L$ as the output of the decoder.

In our decoder, the MH-VolAttn layer learns global dependencies amongst different 3D-volumes, and is calculated as follows:
\begin{align}
&\text{MH-VolAttn}(\mathbf{Y})=\text{cat}(\mathbf{A}^{1},\cdots,\mathbf{A}^{H})\mathbf{W}_{vol} \nonumber \\
&\text{where } \mathbf{A}^{h}=\text{Attn}(\mathbf{Y}\mathbf{W}^h_Q,\mathbf{Y}\mathbf{W}^h_K,\mathbf{Y}\mathbf{W}^h_V.
\end{align}
The MH-ViewVolAttn layer integrates the relevant information across the view and spatial domains, and is calculated as follows:
\begin{align}
&\text{MH-ViewVolAttn}(\mathbf{Y},\mathbf{X}_L)=\text{cat}(\mathbf{A}^{1},\cdots,\mathbf{A}^{H})\mathbf{W} \nonumber \\ 
&\text{where } \mathbf{A}^{h}=\text{Attn}(\mathbf{Y}\mathbf{W}^h_Q,\mathbf{X}_L\mathbf{W}^h_K,\mathbf{X}_L\mathbf{W}^h_V)，
\end{align}
where $\mathbf{W}_{vol}\in\mathbb{R}^{Hd_k\times d}$ and $\mathbf{W}\in\mathbb{R}^{Hd_k\times d}$ are the parameter matrices of the corresponding linear functions. 

Finally, after the 3D-volume decoder, we use a linear function to project the output embeddings of each 3D volume to their 3D output space. Then the predicted 3D volumes are reshaped and grouped to the final reconstruction output.

\section{Experiment}

\subsection{Dataset}

We utilize the ShapeNet dataset \cite{wu20153d} to evaluate the proposed methods and other comparison methods.  We follow 3D-R2N2 \cite{choy20163d} and use the same setting for a fair comparison. Specifically, we use a subset of ShapeNet which consists of 13 categories and 43,783 common 3D objects. For each 3D object, 24 2D-images are rendered from different viewing angles circling around the object.

\subsection{Evaluation Metrics}

\subsubsection{IoU} 
The mean Intersection-over-Union (IoU) calculates the matching degree between predicted 3D voxel grids and their ground-truth grids. A higher IoU value means a better reconstruction result. For each voxel grid, the IoU is defined as:
\begin{align}
\text{IoU} = \frac{\sum_{(i,j,k)}\mathrm{I}(\mathnormal{y}(i,j,k)>t)\mathrm{I}(\mathnormal{\bar{y}}(i,j,k))}{\sum_{(i,j,k)}\mathrm{I}[\mathrm{I}(\mathnormal{y}(i,j,k)>t)+\mathrm{I}(\mathnormal{\bar{y}}(i,j,k))]},
\end{align}
where $y(i,j,k)$ denotes the predicted occupancy probability which is binarized with an optimal fixed voxelization-threshold $t$ for compared methods. $\mathnormal{\bar{y}}(i,j,k)$ is the ground truth at $(i, j, k)$. $\mathrm{I}(\cdot)$ is an indicator function.

\begin{table}
\caption{Parameter sizes of competing methods and pretrained CNNs for 2D-view embeddings in competing methods.}
\label{para}
\begin{tabular}{lll}\hline
             & \begin{tabular}[c]{@{}c@{}}Pretrained CNN used for \\ 2D-view embeddings\end{tabular} & Param. (M) \\\hline
Pix2Vox-A \cite{xie2019pix2vox}    & VGG16 \cite{vgg}          & 114.24     \\
Pix2Vox++/A \cite{xie2020pix2vox++} & ResNet50 \cite{he2016deep}       & 96.31      \\
VolT   & VGG16 \cite{vgg}          & 28.63      \\
VolT+ & 2D-CNN+3D-DCNN & 96.76      \\
EVolT & VGG16 \cite{vgg}          & 29.03 \\\hline    
\end{tabular}
\end{table}

\begin{table*}
\centering
\caption{Comparison of 24-view reconstruction on ShapeNet using IoU and F-Score. The best score for each category is in bold.}
\label{24}
\begin{tabular}{l|lllll|lllll}
\hline
 & \multicolumn{5}{l|}{24-view IoU}                                                                                                                                                                                                                                                      & \multicolumn{5}{l}{24-view F-Score@1\%}                                                                                                                                                                                                                                                  \\ \hline
Category   & \begin{tabular}[c]{@{}l@{}}Pix2Vox-A\\\end{tabular} & \begin{tabular}[c]{@{}l@{}}Pix2Vox++/A\\\end{tabular} & \begin{tabular}[c]{@{}l@{}}VolT\\ 
\end{tabular} & \begin{tabular}[c]{@{}l@{}}VolT+\\ 
\end{tabular} & \begin{tabular}[c]{@{}l@{}}EVolT\\ 
\end{tabular} & \begin{tabular}[c]{@{}l@{}}Pix2Vox-A\\\end{tabular} & \begin{tabular}[c]{@{}l@{}}Pix2Vox++/A\\\end{tabular} & \begin{tabular}[c]{@{}l@{}}VolT\\ 
\end{tabular} & \begin{tabular}[c]{@{}l@{}}VolT+\\ 
\end{tabular} & \begin{tabular}[c]{@{}l@{}}EVolT\\ 
\end{tabular} \\ \hline
airplane   & 0.731     & 0.729       & 0.719      & 0.725       & \textbf{0.741} & 0.635     & 0.614       & 0.604      & 0.618       & \textbf{0.636} \\
bench      & 0.679     & 0.686       & 0.678      & 0.682       & \textbf{0.707} & 0.525     & 0.522       & 0.513      & 0.525       & \textbf{0.548} \\
cabinet    & 0.822     & 0.829       & 0.825      & 0.825       & \textbf{0.832} & 0.448     & 0.456       & 0.452      & 0.455       & \textbf{0.464} \\
car        & 0.880     & 0.883       & 0.884      & 0.885       & \textbf{0.894} & 0.598     & 0.598       & 0.604      & 0.609       & \textbf{0.624} \\
chair      & 0.620     & 0.647       & 0.645      & 0.641       & \textbf{0.681} & 0.318     & 0.341       & 0.339      & 0.340       & \textbf{0.373} \\
display    & 0.599     & 0.613       & 0.635      & 0.613       & \textbf{0.674} & 0.320     & 0.335       & 0.366      & 0.339       & \textbf{0.403} \\
lamp       & 0.475     & 0.493       & 0.478      & 0.481       & \textbf{0.520} & 0.335     & 0.351       & 0.320      & 0.338       & \textbf{0.366} \\
speaker    & 0.751     & 0.762       & 0.762      & 0.753       & \textbf{0.772} & 0.309     & 0.326       & 0.327      & 0.317       & \textbf{0.339} \\
rifle      & 0.676     & 0.686       & 0.663      & 0.693       & \textbf{0.711} & 0.615     & 0.624       & 0.597      & 0.634       & \textbf{0.653} \\
sofa       & 0.764     & 0.782       & 0.781      & 0.776       & \textbf{0.800} & 0.427     & 0.454       & 0.449      & 0.448       & \textbf{0.478} \\
table      & 0.644     & 0.666       & 0.649      & 0.658       & \textbf{0.675} & 0.398     & 0.419       & 0.407      & 0.418       & \textbf{0.431} \\
telephone  & 0.837     & 0.849       & 0.857      & 0.850       & \textbf{0.867} & 0.659     & 0.666       & 0.678      & 0.675       & \textbf{0.687} \\
watercraft & 0.655     & 0.668       & 0.670      & 0.670       & \textbf{0.693} & 0.441     & 0.460       & 0.456      & 0.470       & \textbf{0.494} \\ \hline
Overall    & 0.706     & 0.720       & 0.714      & 0.716       & \textbf{0.738} & 0.462     & 0.473       & 0.468      & 0.475       & \textbf{0.497} \\ \hline
\end{tabular}
\end{table*}

\begin{table*}
\centering
\caption{Comparison of multi-view reconstruction on ShapeNet using IoU and F-Score. The best score for each view number is in bold.}
\label{multi}
\begin{tabular}{lllllllllllll}\hline
\textbf{F-Score@1\%} & 24             & 23             & 22             & 21             & 20             & 18             & 16             & 14             & 12             & 8              & 6              & 4              \\ \hline
3D-R2N2 \cite{choy20163d}         & -              & -              & -              & -              & 0.383          & -              & 0.382          & -              & 0.382          & 0.383          & -              & 0.378          \\
AttSets \cite{yang2020robust}         & -              & -              & -              & -              & 0.448          & -              & 0.447          & -              & 0.445          & 0.444          & -              & 0.430          \\
Pix2Vox-A \cite{xie2019pix2vox}       & 0.462          & 0.462          & 0.462          & 0.462          & 0.462          & 0.461          & 0.461          & 0.461          & 0.460          & 0.458          & 0.456          & 0.452          \\
Pix2Vox++/A \cite{xie2020pix2vox++}      & 0.473             & -              & -              & -              & 0.462          & -              & 0.461          & -              & 0.460          & 0.459          & -              & \textbf{0.457} \\
VolT       & 0.468          & 0.467          & 0.467          & 0.465          & 0.464          & 0.461          & 0.459          & 0.456          & 0.450          & 0.430          & 0.410          & 0.356          \\
VolT+     & 0.475          & 0.475          & 0.474          & 0.474          & 0.474          & 0.473          & 0.472          & 0.471          & 0.469          & \textbf{0.464} & \textbf{0.460} & 0.451          \\
EVolT     & \textbf{0.497} & \textbf{0.496} & \textbf{0.495} & \textbf{0.494} & \textbf{0.492} & \textbf{0.489} & \textbf{0.486} & \textbf{0.481} & \textbf{0.475} & 0.448          & 0.423          & 0.358          \\ \hline
\textbf{IoU}     &                &                &                &                &                &                &                &                &                &                &                &                \\ \hline
3D-R2N2 \cite{choy20163d}         & -              & -              & -              & -              & 0.636          & -              & 0.636          & -              & 0.636          & 0.635          & -              & 0.625          \\
AttSets \cite{yang2020robust}         & 0.694              & -              & -              & -              & 0.693          & -              & 0.692          & -              & 0.688          & 0.685          & -              & 0.675          \\
Pix2Vox-A \cite{xie2019pix2vox}       & 0.706          & 0.706          & 0.706          & 0.706          & 0.706          & 0.705          & 0.705          & 0.705          & 0.704          & 0.702          & 0.700          & 0.697          \\
Pix2Vox++/A \cite{xie2020pix2vox++}     & 0.720              & -              & -              & -              & 0.719          & -              & 0.718          & -              & 0.717          & \textbf{0.715} & -              & \textbf{0.708} \\
VolT       & 0.714          & 0.713          & 0.712          & 0.711          & 0.711          & 0.708          & 0.706          & 0.703          & 0.699          & 0.681          & 0.662          & 0.605          \\
VolT+      & 0.716          & 0.716          & 0.716          & 0.715          & 0.715          & 0.714          & 0.714          & 0.713          & 0.711          & 0.707          & 0.704          & 0.695          \\
EVolT     & \textbf{0.738} & \textbf{0.738} & \textbf{0.737} & \textbf{0.735} & \textbf{0.735} & \textbf{0.732} & \textbf{0.729} & \textbf{0.726} & \textbf{0.720} & 0.698          & 0.675          & 0.609  \\ \hline       
\end{tabular}
\end{table*}

\begin{figure*}
	\centering
	\includegraphics[width=1\linewidth]{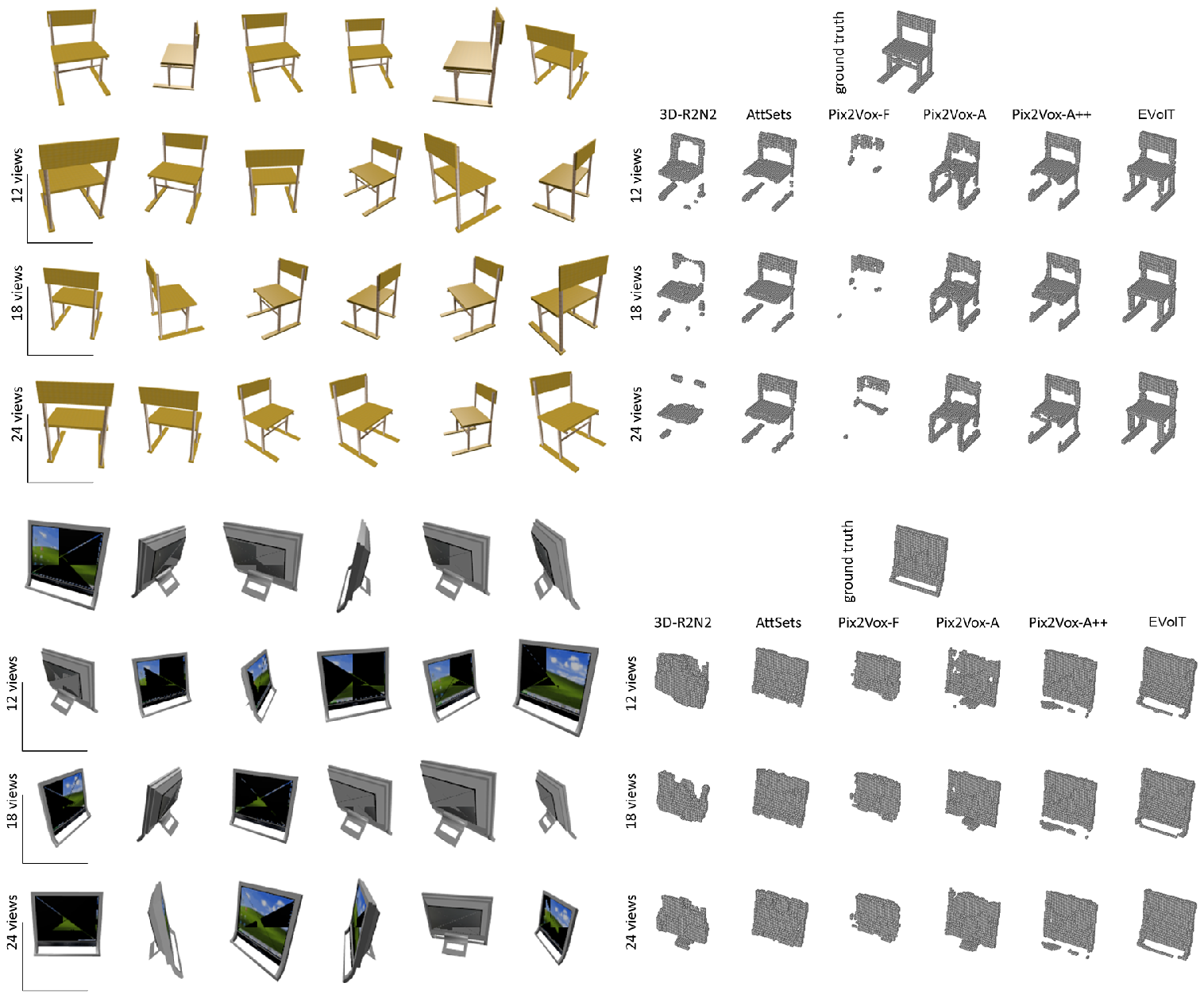}
	\centering
	\caption{Qualitative 3D object reconstruction results on ShapeNet based on different number of input 2D-view images.}
	\label{fig:qual}
\end{figure*}

\subsubsection{F-Score} 

Compared with IoU, F-score \cite{Tatarchenko_2019_CVPR, xie2020pix2vox++} explicitly evaluates the distance between object surfaces, which is more interpretable. F-score is formally defined as the harmonic mean between precision $\mathrm{P}(d)$ and recall $\mathrm{R}(d)$ with a distance threshold $d$:
\begin{align}
\text{F-Score}(d) = \frac{2\mathrm{P}(d)\mathrm{R}(d)}{\mathrm{P}(d)+\mathrm{R}(d)} 
\end{align}
A higher F-score with a stringent distance threshold indicates a better reconstruction result. 

In F-Score, $\mathrm{P}(d)$ estimates the reconstruction accuracy by counting the portion of reconstructed points lying within the distance $d=1\%$ to the ground truth. $\mathrm{R}(d)$ quantifies the reconstruction completeness by counting the percentage of ground-truth points lying within the distance $d$ to the reconstruction. These two metrics are defined as follows:
\begin{align}
&\mathrm{P}(d) = \frac{1}{|\mathcal{R}|}\sum_{\mathbf{r}\in\mathcal{R}}[\mathnormal{e}_{\mathbf{r}\rightarrow\mathcal{G}}<d], \mathnormal{e}_{\mathbf{r}\rightarrow\mathcal{G}}=\min_{\mathbf{r}\in\mathcal{G}}\|\mathbf{r}-\mathbf{g}\|\\
&\mathrm{R}(d) = \frac{1}{|\mathcal{G}|}\sum_{\mathbf{g}\in\mathcal{G}}[\mathnormal{e}_{\mathbf{g}\rightarrow\mathcal{R}}<d], \mathnormal{e}_{\mathbf{g}\rightarrow\mathcal{R}}=\min_{\mathbf{r}\in\mathcal{R}}\|\mathbf{g}-\mathbf{r}\|,
\end{align}
where $[\cdot]$ is the Iverson bracket.
$\mathcal{G}$ is the ground-truth point set and $\mathcal{R}$ is the reconstructed point set being evaluated.
We apply F-Score with the same setting in \cite{xie2020pix2vox++}.

\subsubsection{Divergence measurement for multi-view representations}\label{dis}

We also define a metric to explore the convergence of multi-view representations in different layers. Since the convergence has a positive correlation with the divergence decay of multi-view attentions, we utilize a similarity measure based on multi-view attentions to evaluate the divergence enhancing ability in our method.

In each view attention layer, an attention-score matrix $\mathbf{S}=\text{softmax}(\frac{\mathbf{Q}\mathbf{K}^T}{\sqrt{d_{k}}})$ contains view-to-view attention vectors. The $m$-th row of $\mathbf{S}$, denoted as $\mathbf{s}_{m}$, is an attention-score vector where each element represents its attention weight to another view. For 3D reconstruction of a specific object, the Euclidean distance measuring the similarity of multi-view attentions, is defined as
\begin{align}\label{Dis}
\begin{split}
&\mathnormal{D}=\frac{1}{N_{view}}\sum_m\|\mathbf{s}_{m}-\bar{\mathbf{s}}\|_2\\
&\text{where} \quad\mathbf{\bar{s}}=\frac{1}{N_{view}}\sum_m\mathbf{s}_{m}.
\end{split}
\end{align}
Here, a small $D$ means a more considerable similarity and the convergence of multi-view representations.

\subsection{Implementation Details}

We set the batch size to 64 and the view image size to $224 \times 224$ for training. The 3D spatial size of the voxelized output is set to $32\times32\times 32$. The VolT and its two variants VolT+, and EVolT are trained by an AdamW optimizer \cite{adamw} with a $\beta_{1}$ of 0.9 and a $\beta_{2}$ of 0.999.

Table \ref{para} shows the parameter sizes of competing methods and pretrained CNNs for the initial view embeddings used in different competing methods.  Compared with Pix2Vox-A \cite{xie2019pix2vox} and Pix2Vox++/A \cite{xie2020pix2vox++},  the parameter size of EVolT is only around $30\%$ of them. To obtain the reported best results, Pix2Vox-A and Pix2Vox++/A both adopt an additional 3D-CNN-based refiner containing another 3D-CNN and 3D-DCNN.
In contrast, our proposed end-to-end methods do not need additional refiner and can also achieve the best results. To testify the effect of the transformer architecture, in VolT+, we apply an advanced CNN feature extraction model for 2D-view embeddings from the 2D-CNN and 3D-DCNN without the last layer in Pix2Vox-A.


\begin{figure}
	\centering
	\includegraphics[width=1\linewidth]{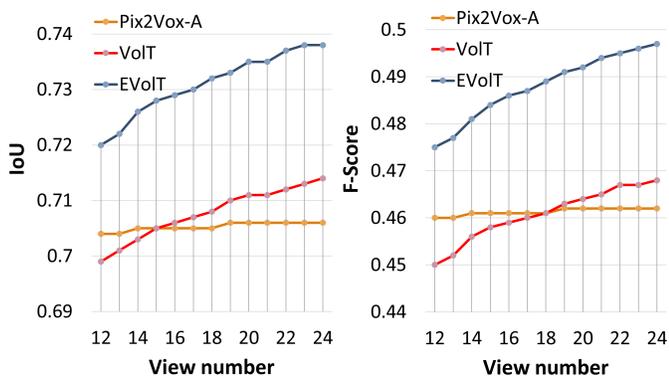}
	\centering
	\caption{Effect of the view-divergence enhancing function on 3D reconstruction results.}
	\label{fig:ablat_3d}
\end{figure}

\begin{figure*}
	\centering
	\includegraphics[width=1\linewidth]{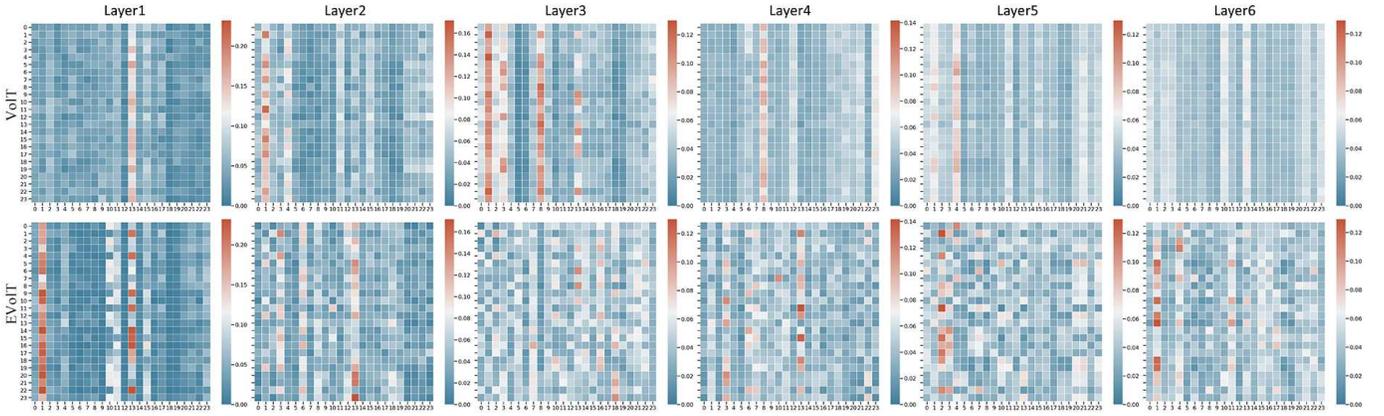}
	\centering
	\caption{Multi-view attention-matrix visualization in VolT and EVolT.}
	\label{fig:ablat_attn}
\end{figure*}

\begin{figure}
	\centering
	\includegraphics[width=0.8\linewidth]{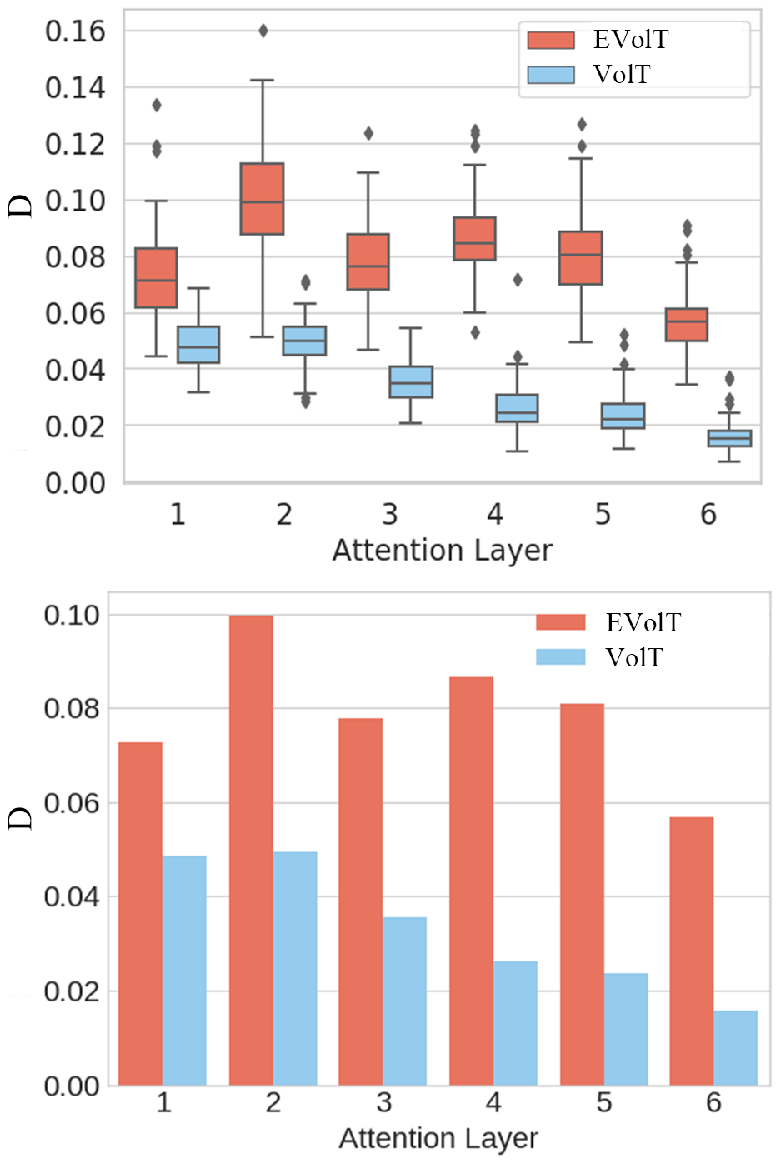}
	\centering
\caption{Discrepancy amongst multi-view representations in VolT and EVolT.}
	\label{fig:attn_div}
\end{figure}

\subsection{Multi-view 3D Object Reconstruction}

\subsubsection{\textbf{Quantitative results}}

Here, we show the quantitative results of compared methods on ShapeNet using different evaluation metrics. Table \ref{24} shows the comparison of 24-view object reconstruction on ShapeNet using IoU and F-Score metrics. The highest value for each category is highlighted in bold. This table shows that EVolT reaches the highest IoU and F-score amongst the compared methods. VolT gets moderate results between Pix2Vox-A and Pix2Vox++/A. VolT+ works better than VolT because it uses better initial features. However, VolT+ still falls behind EVolT even the EVolT is simply based on the plain VGG features. These observations indicate that the view-divergence enhancing function in EVolT plays an indispensable role in increasing its performance against the compared methods.

Table \ref{multi} shows the multi-view object reconstruction results on ShapeNet. 
The best score for each number of views is highlighted in bold.
This table shows that the performances of our methods increase appreciably as the number of views increases. In comparison, other compared methods increase slightly when the view number enlarges. For example, the mean IoU of EVolT increases by 0.04 from 8 views to 24 views, which is eight times the improvement of Pix2Vox++/A. This observation indicates that the proposed Transformer-based methods has better scaling ability and can learn a more comprehensive 3D representation with the increase of view number. We can also see from this table that our proposed methods get the best  F-Score when the view number is larger than 6 and get the best IoU when the view number is higher than 12.


\subsubsection{\textbf{Qualitative results}}

In Figure \ref{fig:qual}, we show the qualitative results of 3D object reconstruction of different methods on ShapeNet. 
In each object sample, we provide object reconstruction results from different number of input views, i.e., 12 views, 18 views, and 24 views.
The first two rows on the left part of Figure \ref{fig:qual} show the 12 input views of an object, and the corresponding reconstruction results of competing methods are shown at the second row on the right.
Similarly, the first three rows on the left part are the 18 input views corresponding to the results on the right.

Figure \ref{fig:qual} shows that EVolT can obtain more accurate and complete 3D reconstruction against compared methods. For example, the EVolT results in the last column successfully recover chair legs and monitor stand while other methods only show incomplete parts. More qualitative results can be found in our Supplementary Material.

\subsection{Ablation Study}

The following ablation experiments are made to verify the effectiveness of the proposed view-divergence enhancing function.

\subsubsection{Effect on 3D reconstruction accuracy}

In Figure \ref{fig:ablat_3d}, we quantitatively evaluate the influence of the view-divergence enhancing function on 3D reconstruction results by comparing EVolT with VolT and Pix2Vox-A. From Figure \ref{fig:ablat_3d}, we can observe that EVolT significantly outperforms VolT that achieves better results than Pix2Vox-A. This indicates the positive effect of the view-divergence enhancing function on 3D reconstruction results.

\subsubsection{Effect on the view divergence}

In Figure \ref{fig:ablat_attn}, we visualize the view-to-view attention matrix in different layers by VolT and EVolT. We set the input view number to 24 in this experiment. In the attention matrix at each layer, the m-th row shows an attention vector where each element is the attention weight of the m-th view to another view.  From the top of Figure \ref{fig:ablat_attn}, we can observe that rows become more similar in a standard transformer as the attention layers go deeper. As a comparison, in EVolT, we can see the diversity of multi-view attention still keeps in deep layers, which means that the divergence enhancement function in EVolT can effectively slow down the convergence degradation of multi-views in deeper layers. 

In Figure \ref{fig:attn_div}, the similarity measurement score $D$ in Eq. \ref{Dis} is also recorded to analyze the convergence amongst multi-view representations in each layer. For 100 randomly chosen objects, we plot $D$ in different layers displayed in the left of Figure \ref{fig:attn_div}, and the right shows the average values of $D$. A small $D$ suggests a significant convergence amongst multi-view representations. As shown in Figure \ref{fig:attn_div}, the value of $D$ obtained by VolT declines gradually with deepening the 2D-view encoder layer while the value of $D$ of EVolT at the same layer keeps higher than that of VolT.

The ablation studies indicate that the view-divergence enhancing function plays an essential role in improving the proposed EVolT performance and relieving the convergence amongst multi-view representations in different layers.

\section{Conclusion and Future Works} 
\label{sec:conclusion}

In this paper, we propose a Transformer-based framework for multi-view 3D reconstruction and achieves state-of-the-art accuracy on ShapeNet with fewer parameters than other CNN-based methods. We propose three versions of the method (VolT, VolT+ and EVolT) to explore view and spatial domain relationships for multi-view 3D reconstruction. Meantime, we explore the problem of divergence decay for the multi-view information in deeper layers and proposed view-divergence enhancing function to ease such a problem. In our future work, we will work on exploring the interpretability of the proposed framework and give its explanation for 3D reconstruction. We also plan to build an interpretable way to visualize and understand the correspondence between the latent representation and multiple input views.


{\small
\bibliographystyle{IEEEtran}
\bibliography{ref}
}

\newpage
\onecolumn

\begin{center}
	\huge{Supplementary Material: \\Multi-view 3D Reconstruction with Transformer}
\end{center}

\setcounter{section}{0}
\section{Comparison of Structures in Competing Methods}
\begin{figure*}[htp]
	\centering
	\includegraphics[width=.8\linewidth]{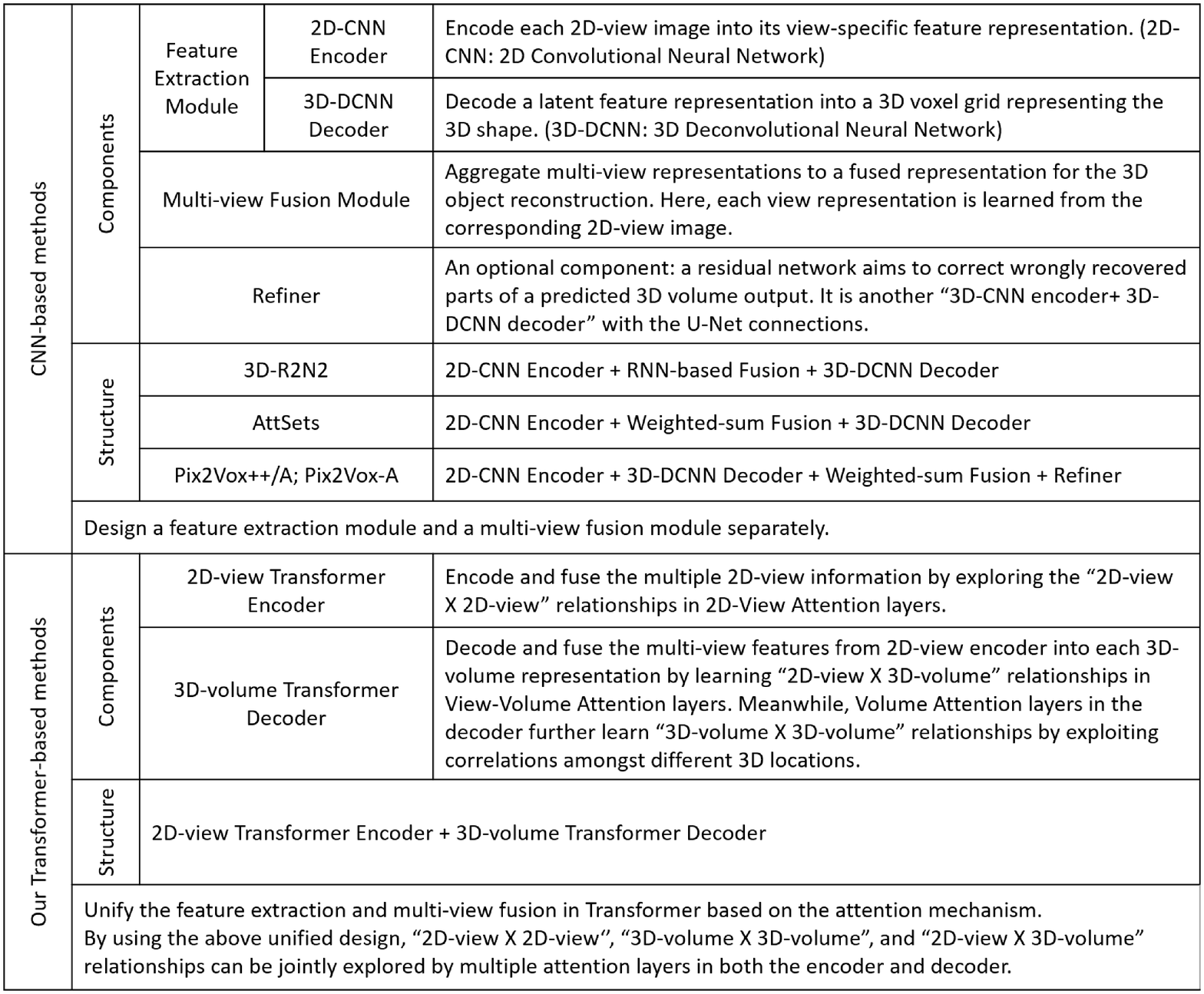}
	\centering
	\caption{Comparison of CNN-based methods and the proposed Transformer-based methods.}
	\label{fig:vs}
\end{figure*}

Table \ref{fig:vs} gives a comparison between the proposed transformer-based 3D reconstruction methods and existing CNN-based methods from the perspective of components and structure.

\begin{figure*}[ht!]
	\setlength\tabcolsep{1.5pt}
	\centering
	\begin{tabular}{ccc}
		Layer 1 & Layer 2 & Layer 3 \\
		\includegraphics[width=0.32\textwidth]{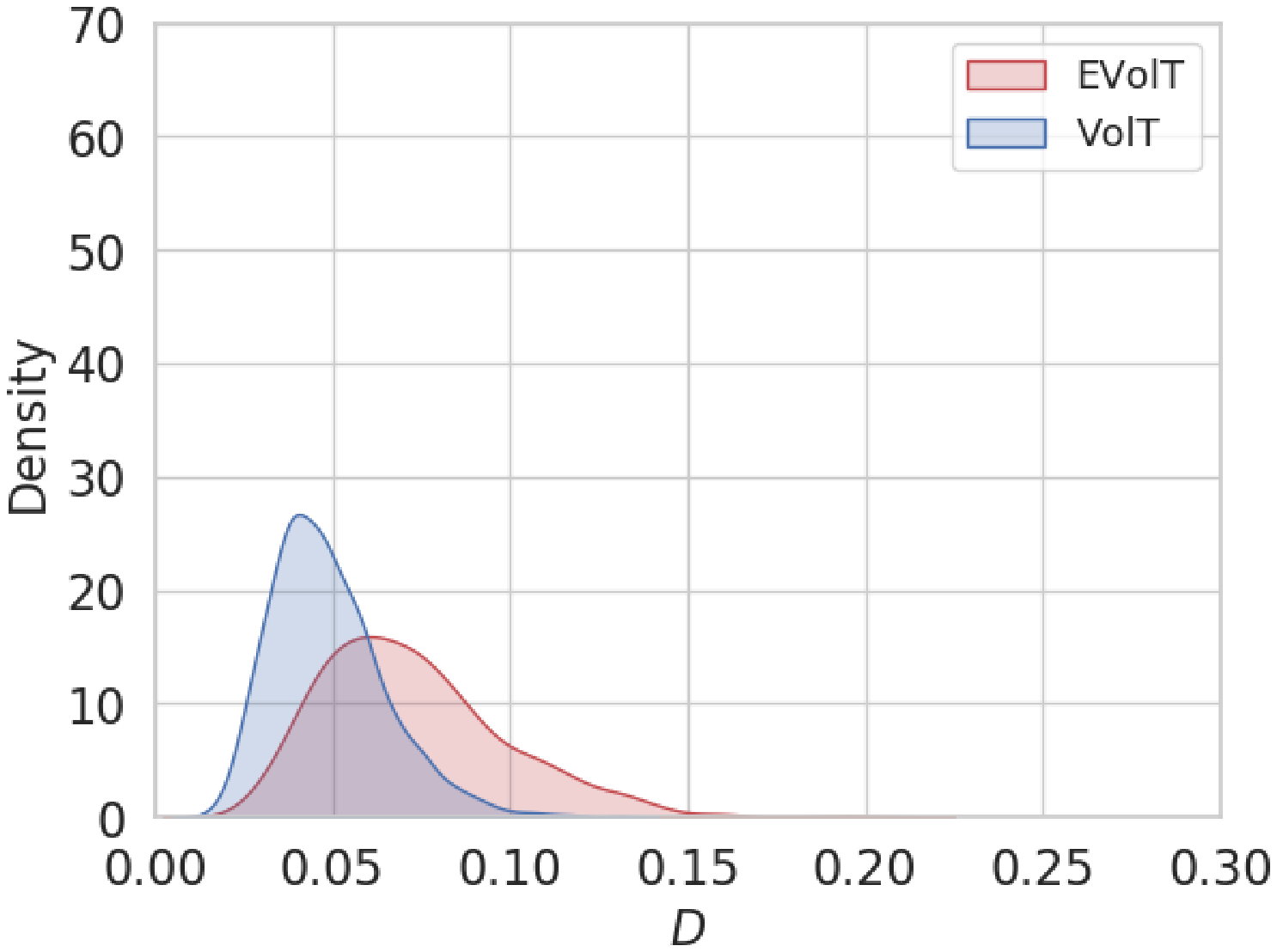} &
		\includegraphics[width=0.32\textwidth]{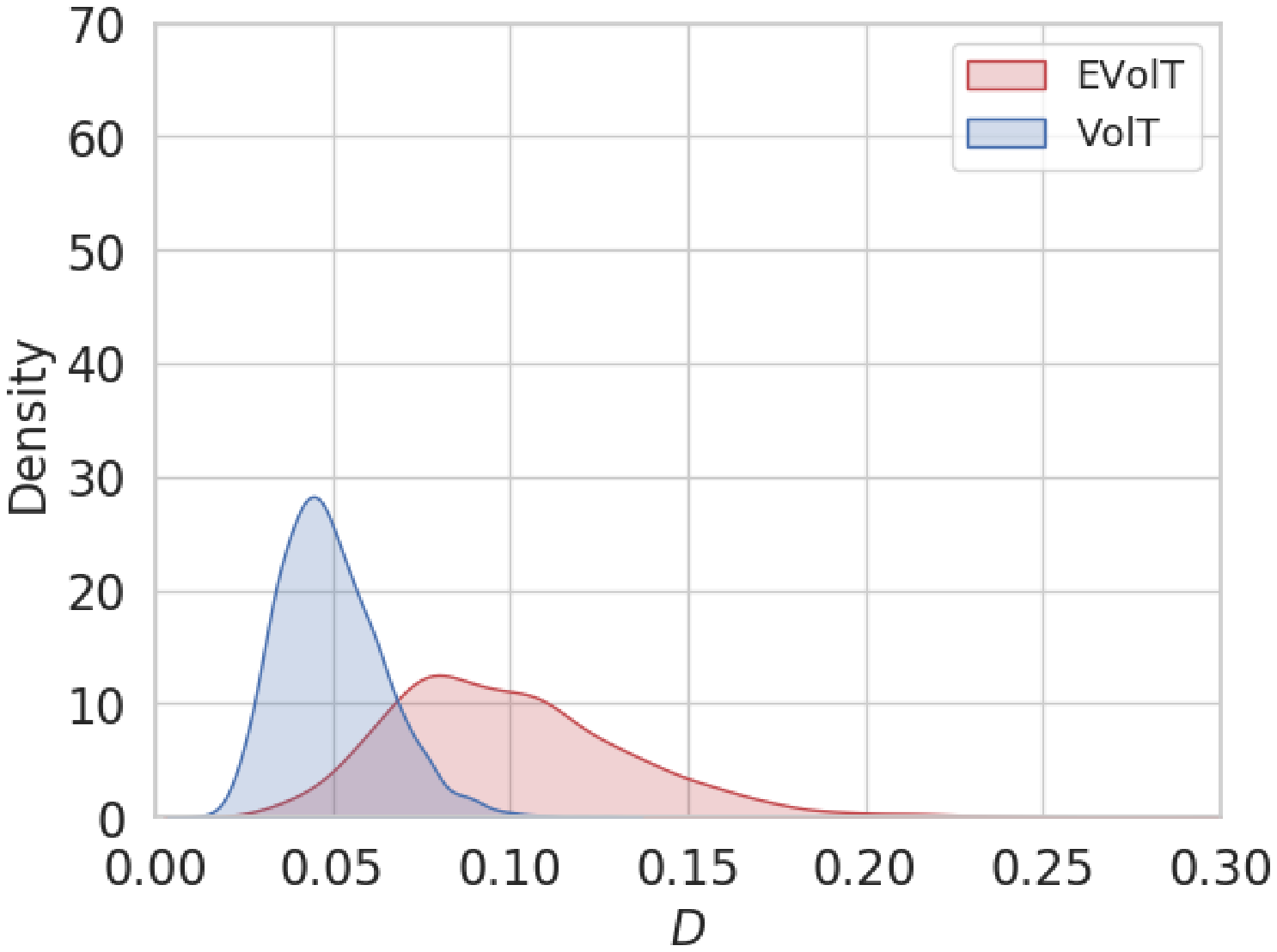} &
		\includegraphics[width=0.32\textwidth]{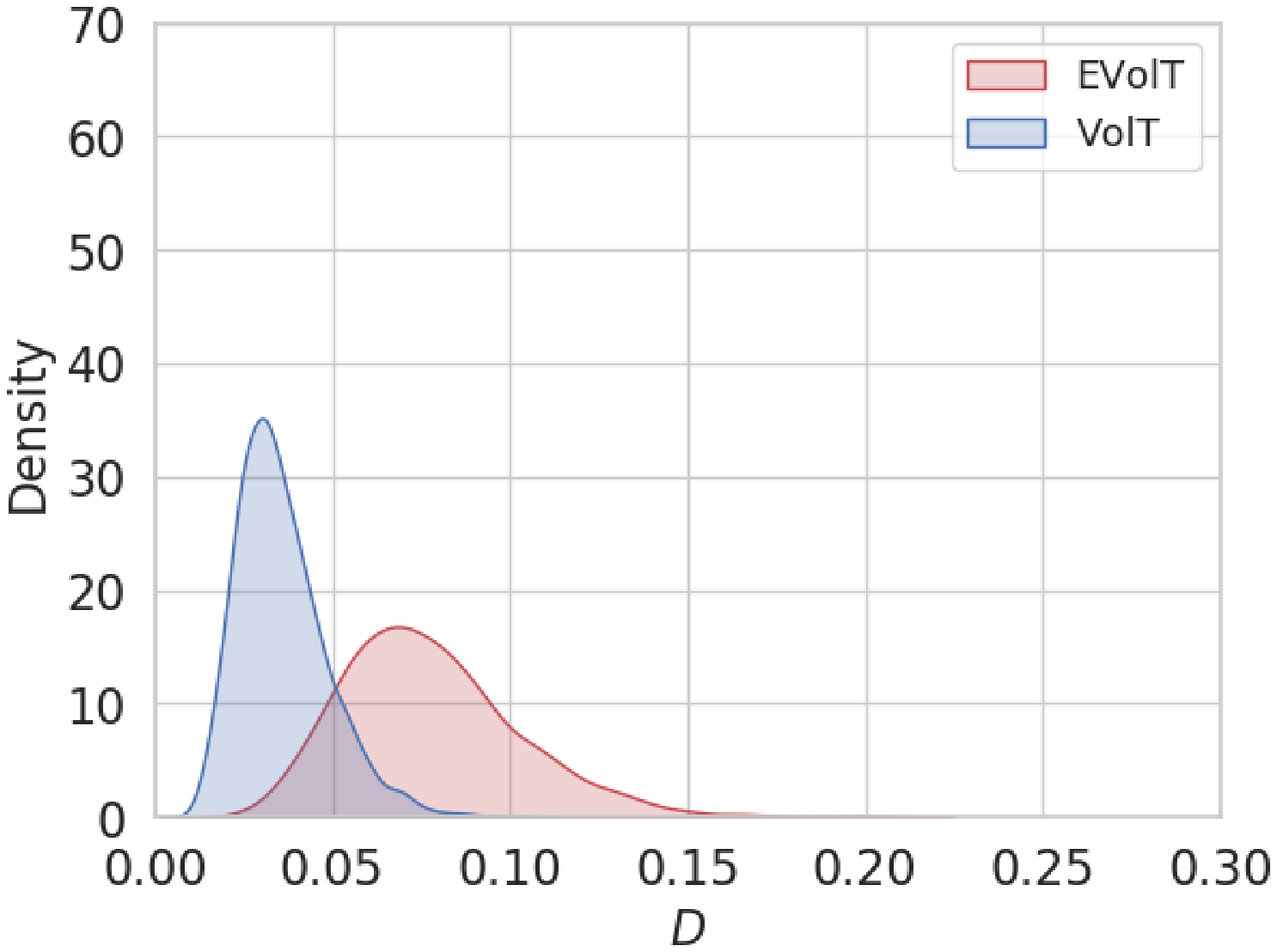}\\
		Layer 4 & Layer 5 & Layer 6 \\
		\includegraphics[width=0.32\textwidth]{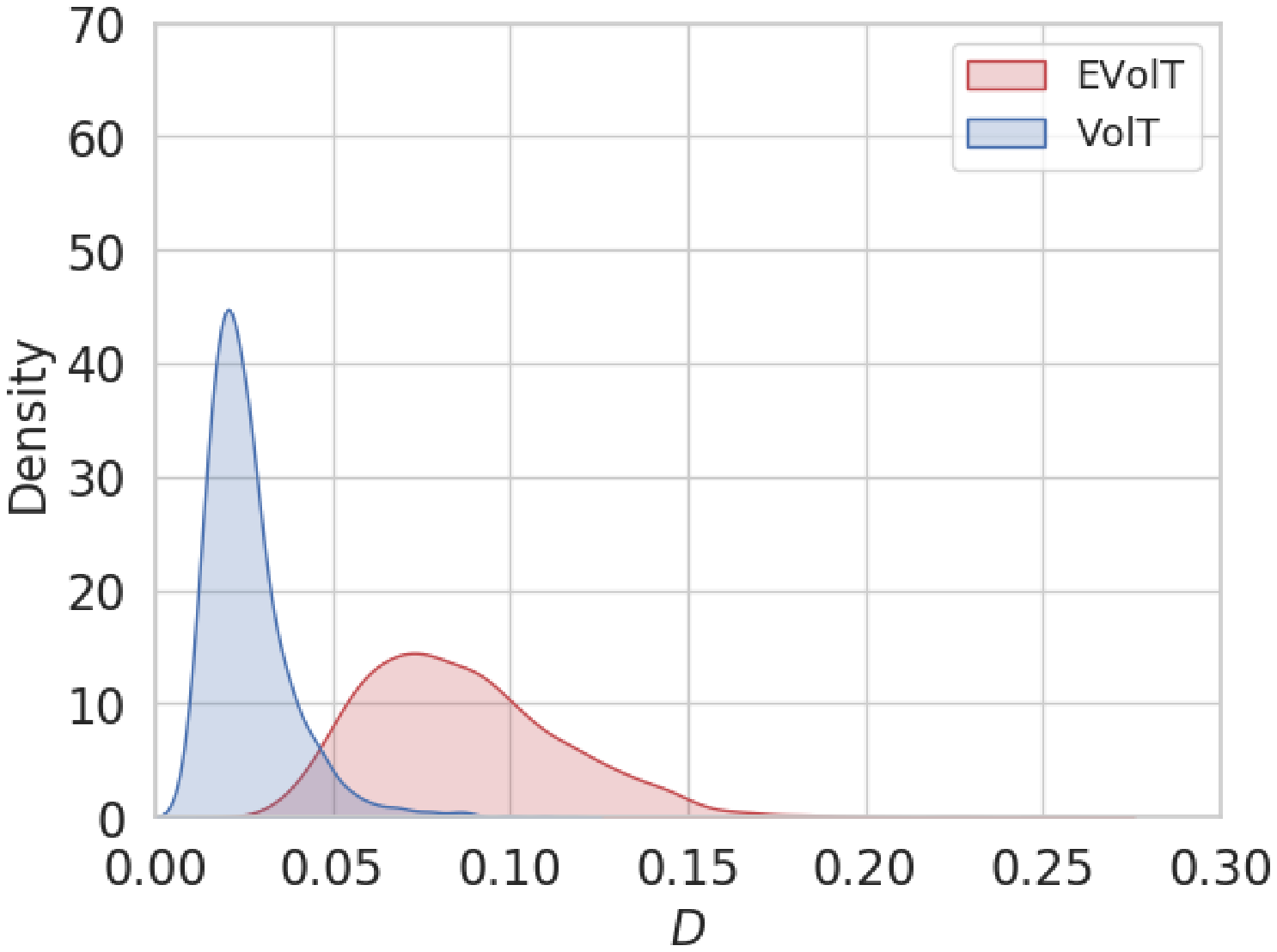} &
		\includegraphics[width=0.32\textwidth]{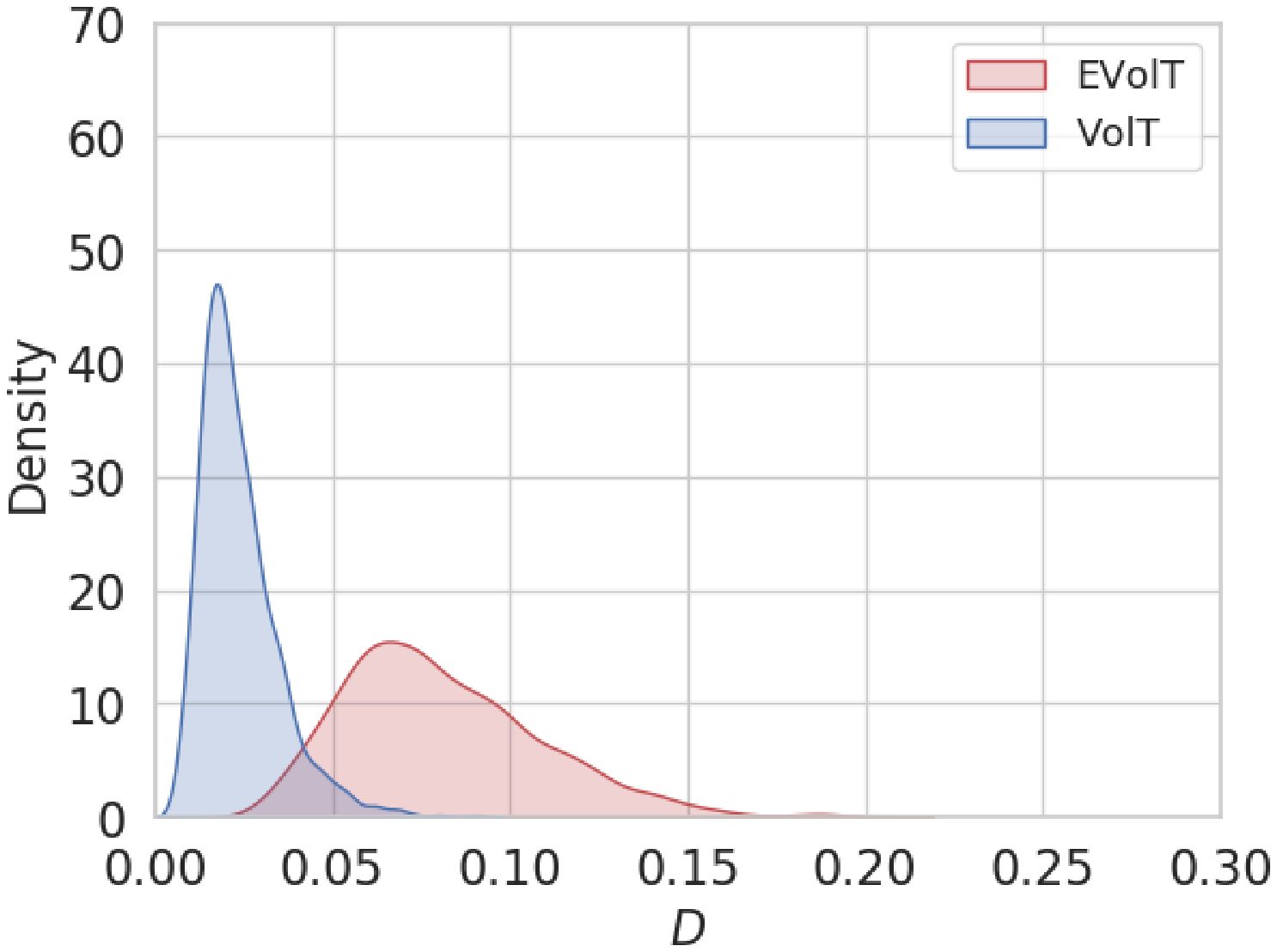} &
		\includegraphics[width=0.32\textwidth]{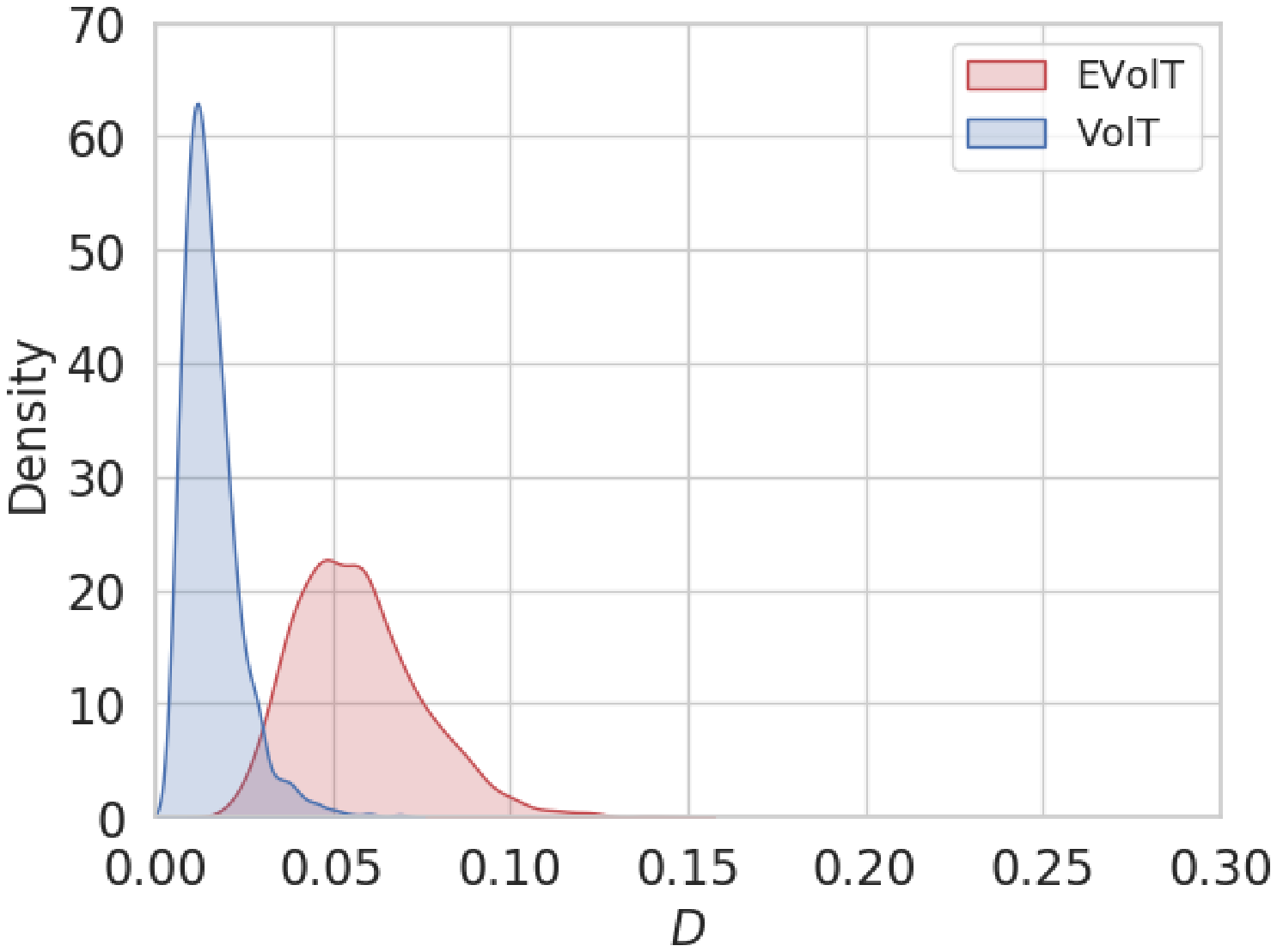} 
	\end{tabular}
	\caption{Kernel density estimation of $D$ value in different attention layers for VolT and EVolT.}
	\vspace{-0.05in}
	\label{fig:denstiy}
\end{figure*}

\section{Additional Results}

\subsection{View Divergence}

In Figure \ref{fig:denstiy}, we plot the estimated probability density of the $D$ value at different attention layers for VolT and EVolT. We use kernel density estimation (KDE) to compute the probability density and explore the convergence of multi-view representations in different attention layers. 
A small $D$ means a more considerable convergence of multi-view representations.

In each view attention layer, the probability density function $\hat{p}(\mathnormal{D})$ of $D$ is estimated as
\begin{align}\label{D}
\begin{split}
&\hat{p}(\mathnormal{D})=\frac{1}{N_{object}N_{view}h}\sum_i^{N_{object}}\sum_m^{N_{view}}K(\frac{\mathnormal{D}_m^i-\mathnormal{D}}{h})\\
&\text{where} \quad\mathnormal{D}_m^i=\| \mathbf{s}_{m}^i-\frac{1}{N_{view}}\sum_m^{N_{view}}\mathbf{s}_{m}^i \|_2
\end{split}
\end{align}
where $\mathbf{s}_{m}^i$ is the attention vector of the $m$-th view for the $i$-th object. The number of random objects is set to $N_{object}=100$. The input view number is set to $N_{view}=24$.
Here, we used the Gaussian kernel $K(x)=\frac{1}{\sqrt{2\pi}}\exp(-\frac{x^{2}}{2})$.
$h$ is computed by the rule of thumb of Scott.

It is shown in Figure \ref{fig:denstiy} that the density of EVolT has a much larger variance than that of the VolT. Also, as the attention layers go deeper, the $D$ value of the VolT gradually moves closer to 0 while the EVolT can still cover a larger range of $D$ values. This indicates that the divergence enhancement function in EVolT can effectively slow down the convergence degradation of multi-views in deeper layers.

\subsection{Qualitative Results}
\begin{figure*}[htp]
	\begin{center}
		\includegraphics[width=.98\linewidth]{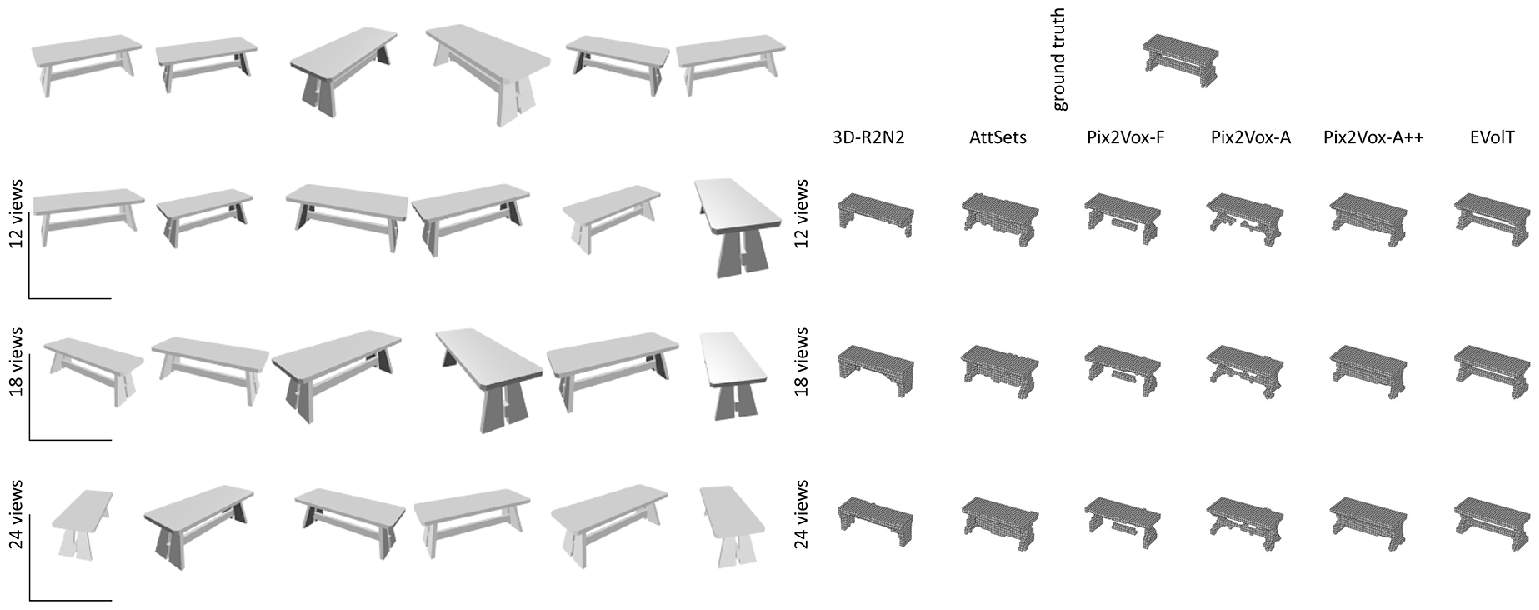} \\
		\includegraphics[width=.98\linewidth]{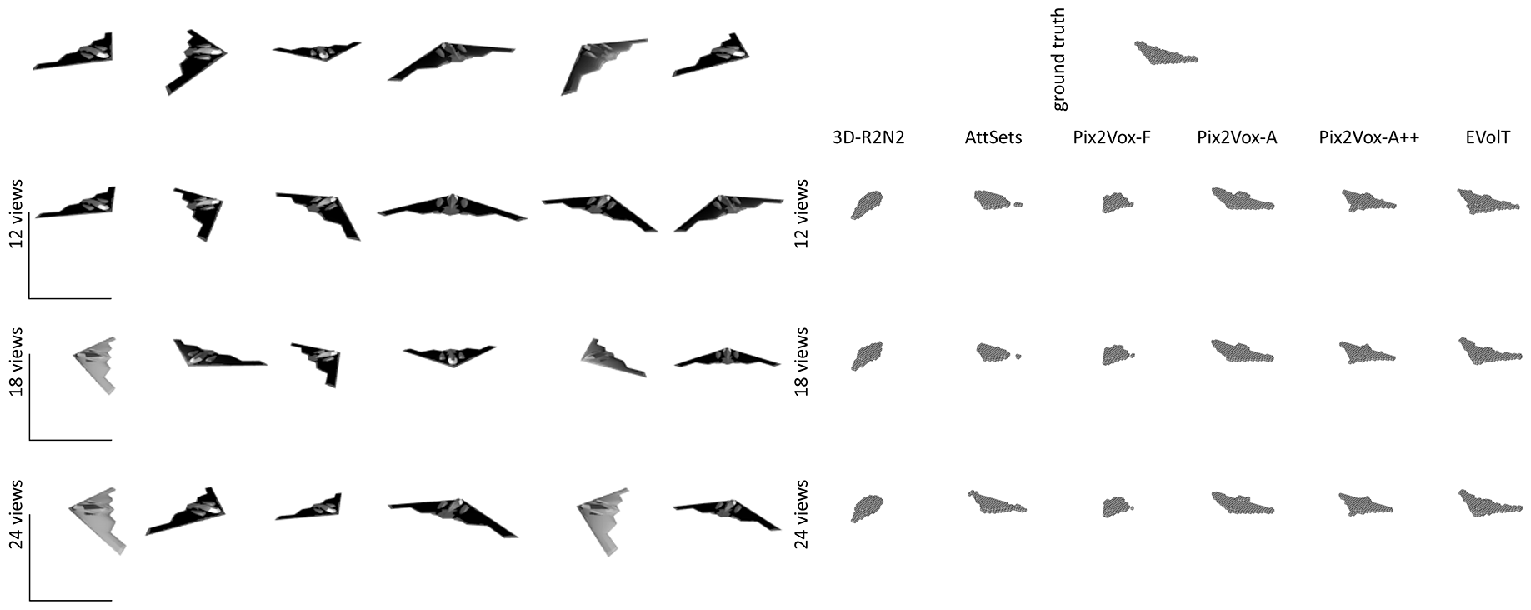} \\
		\includegraphics[width=.98\linewidth]{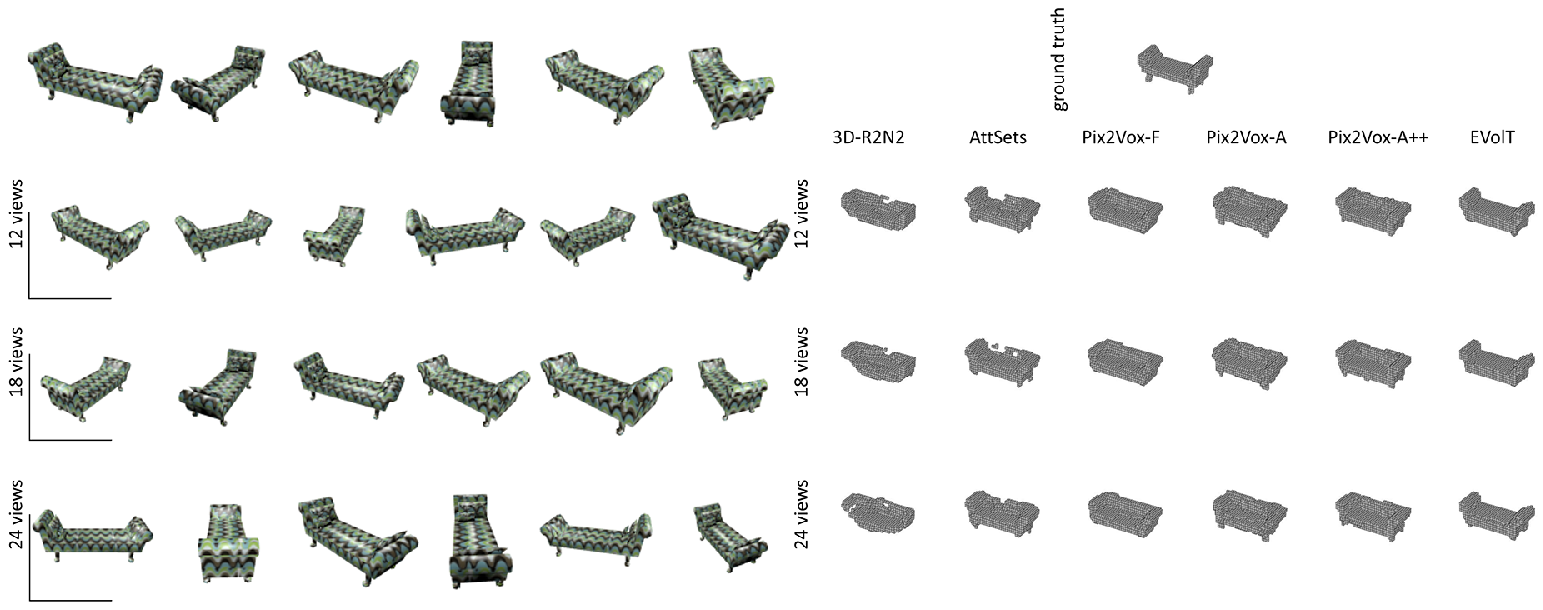} 
	\end{center}
	\caption{Qualitative reconstruction results of competing methods for bench (top), aeroplane (middle), and sofa (bottom).}
	\vspace{-0.05in}
	\label{fig:qual1}
\end{figure*}

\begin{figure*}[htp]
	\begin{center}
		\includegraphics[width=.98\linewidth]{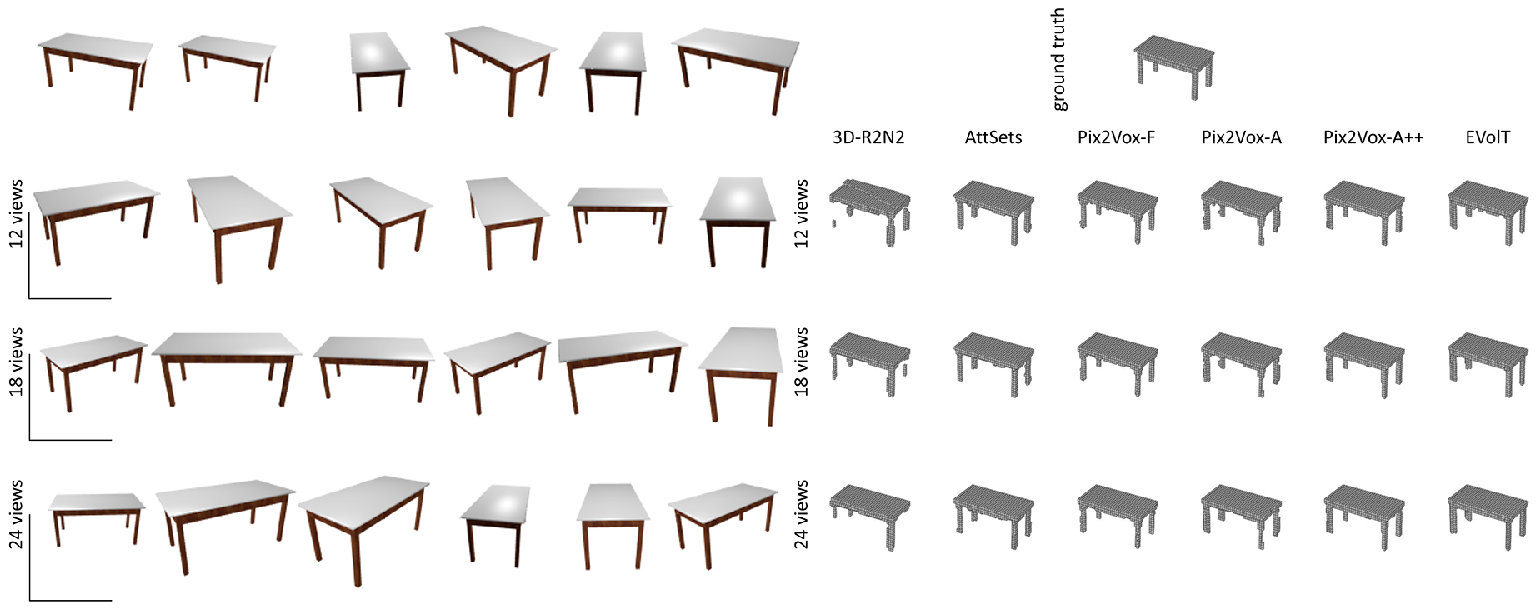} \\
		\includegraphics[width=.98\linewidth]{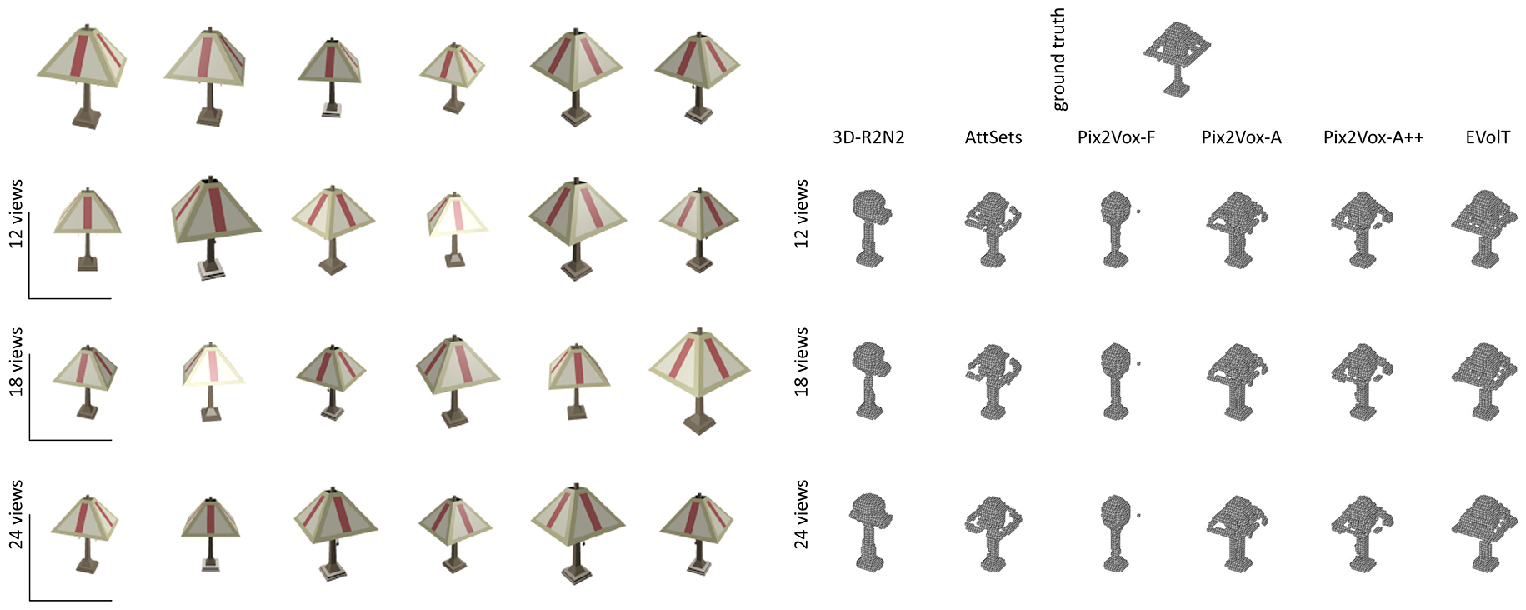} \\
		\includegraphics[width=.98\linewidth]{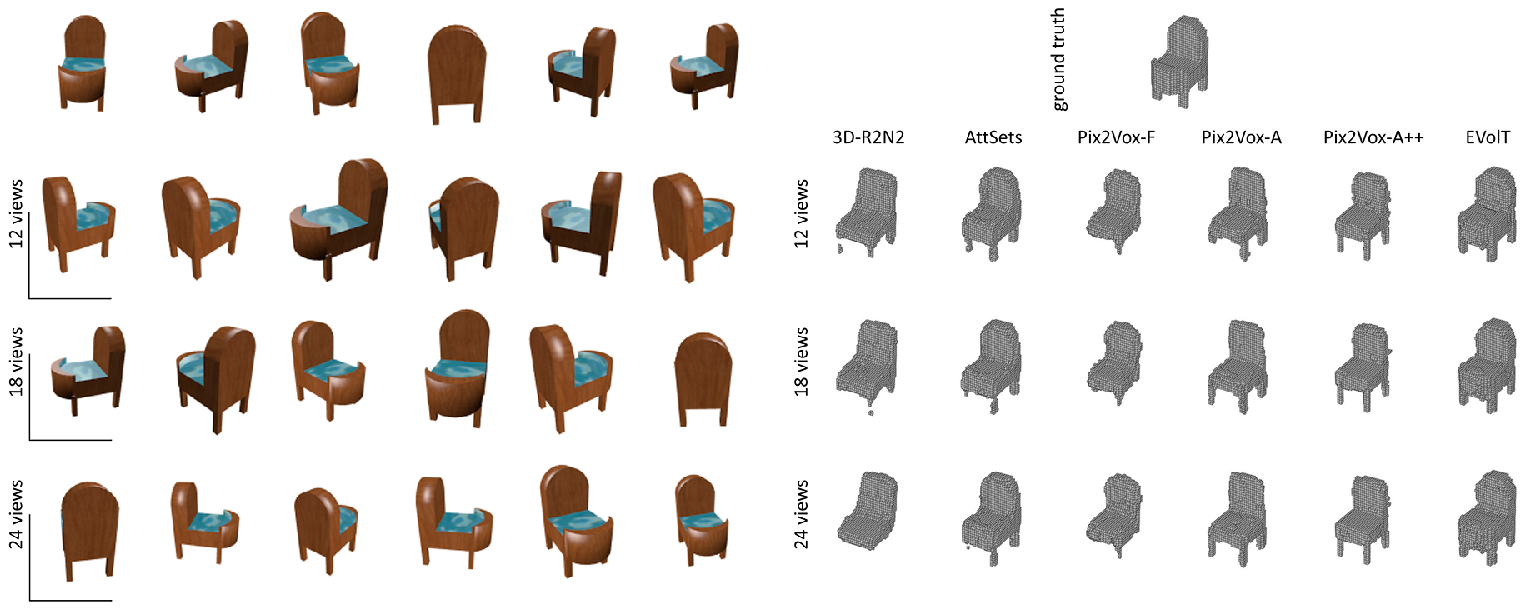} 
	\end{center}
	\caption{Qualitative reconstruction results of competing methods for table (top), lamp (middle), and chair (bottom).}
	\vspace{-0.05in}
	\label{fig:qual2}
\end{figure*}

\begin{figure*}[htp]
	\begin{center}
		\includegraphics[width=.98\linewidth]{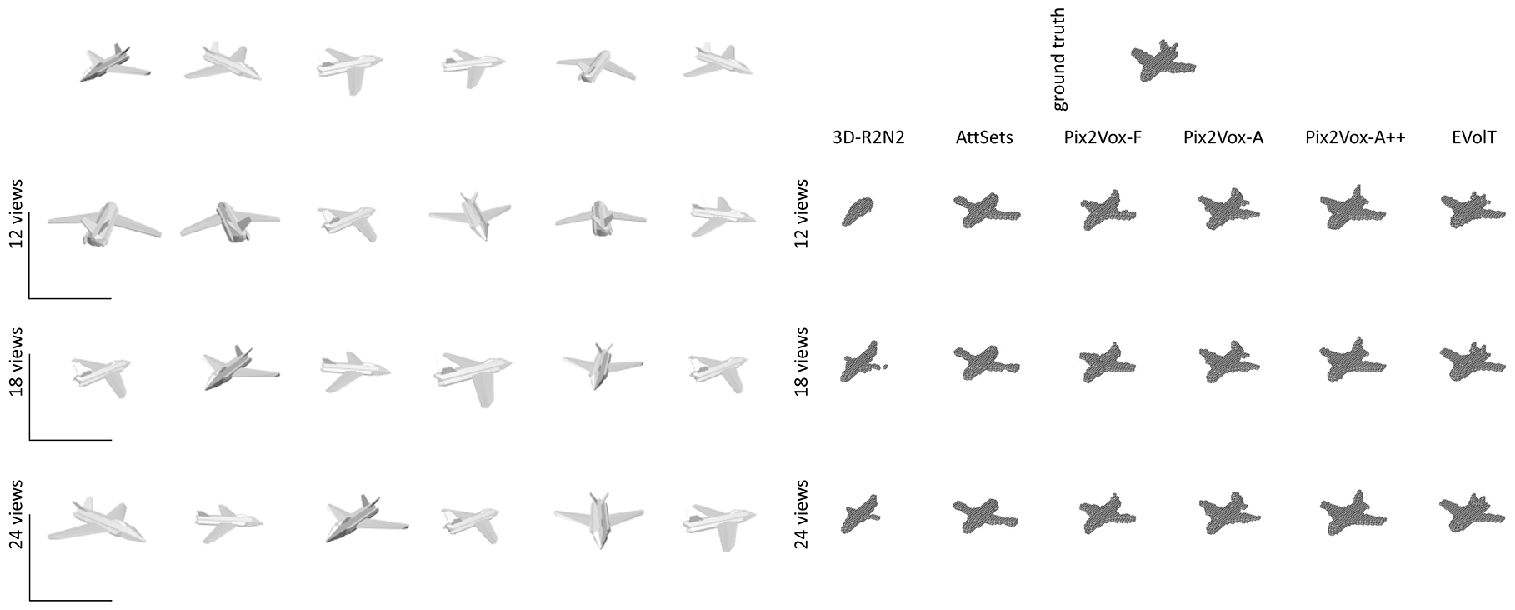} \\
		\includegraphics[width=.98\linewidth]{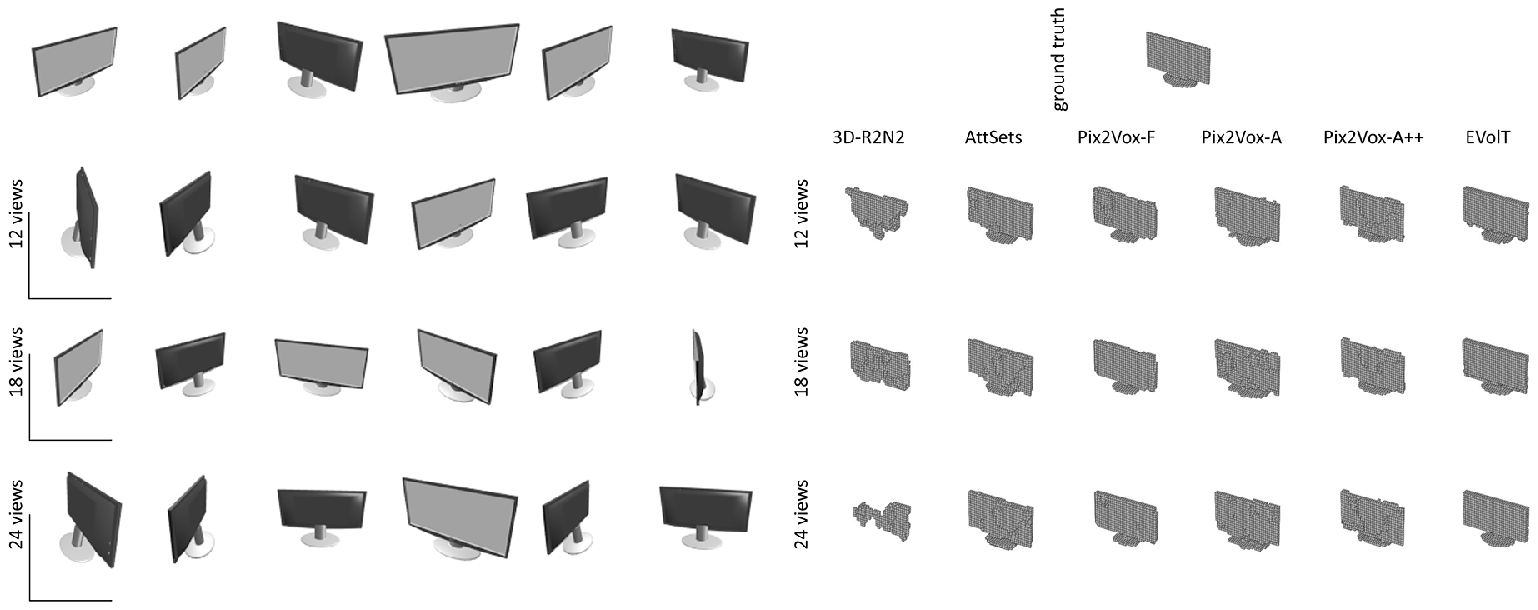} \\
		\includegraphics[width=.98\linewidth]{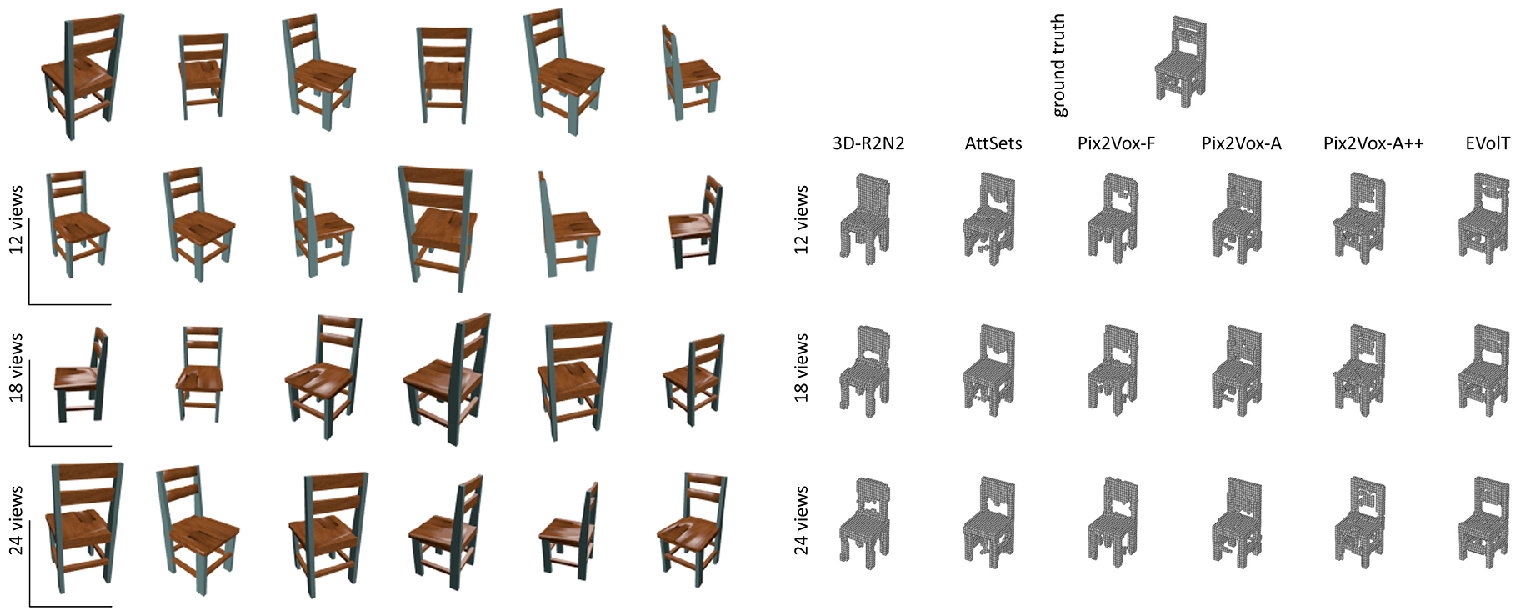} 
	\end{center}
	\caption{Qualitative reconstruction results of competing methods for aeroplane (top), display (middle), and chair (bottom).}
	\vspace{-0.05in}
	\label{fig:qual3}
\end{figure*}

\begin{figure*}[htp]
	\begin{center}
		\includegraphics[width=.98\linewidth]{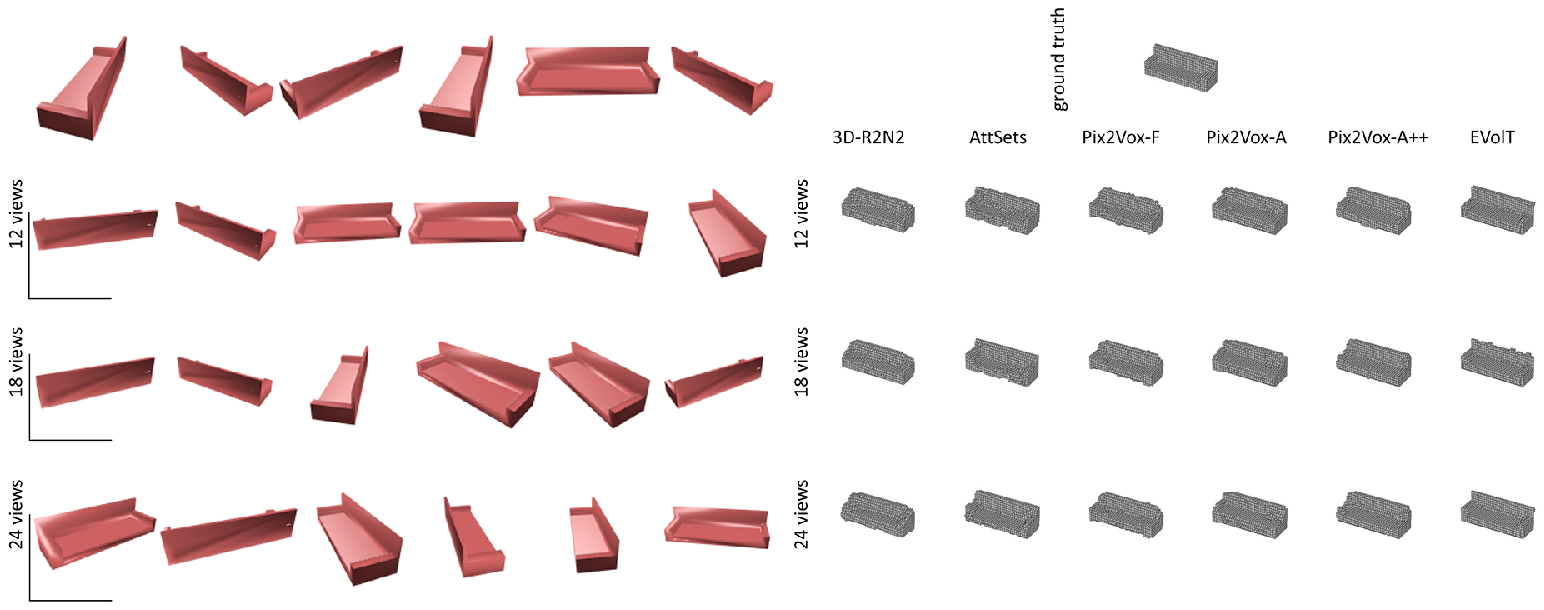} \\
		\includegraphics[width=.98\linewidth]{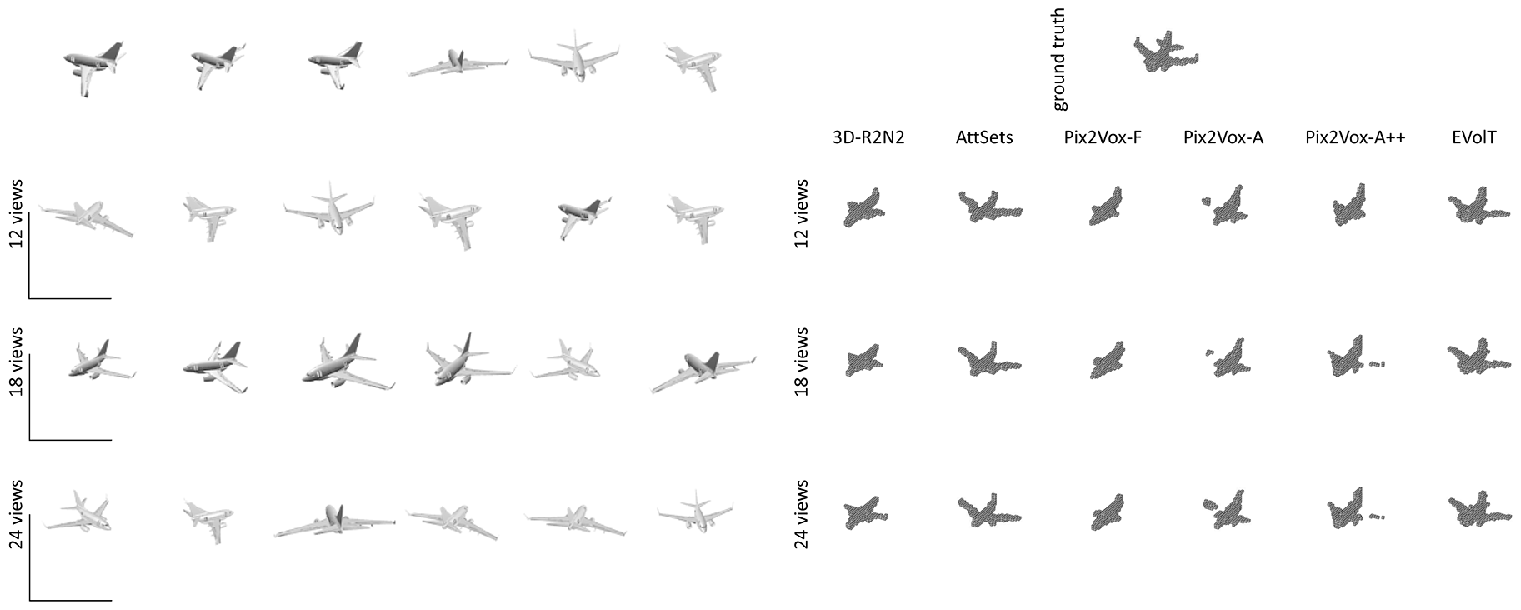} \\
		\includegraphics[width=.98\linewidth]{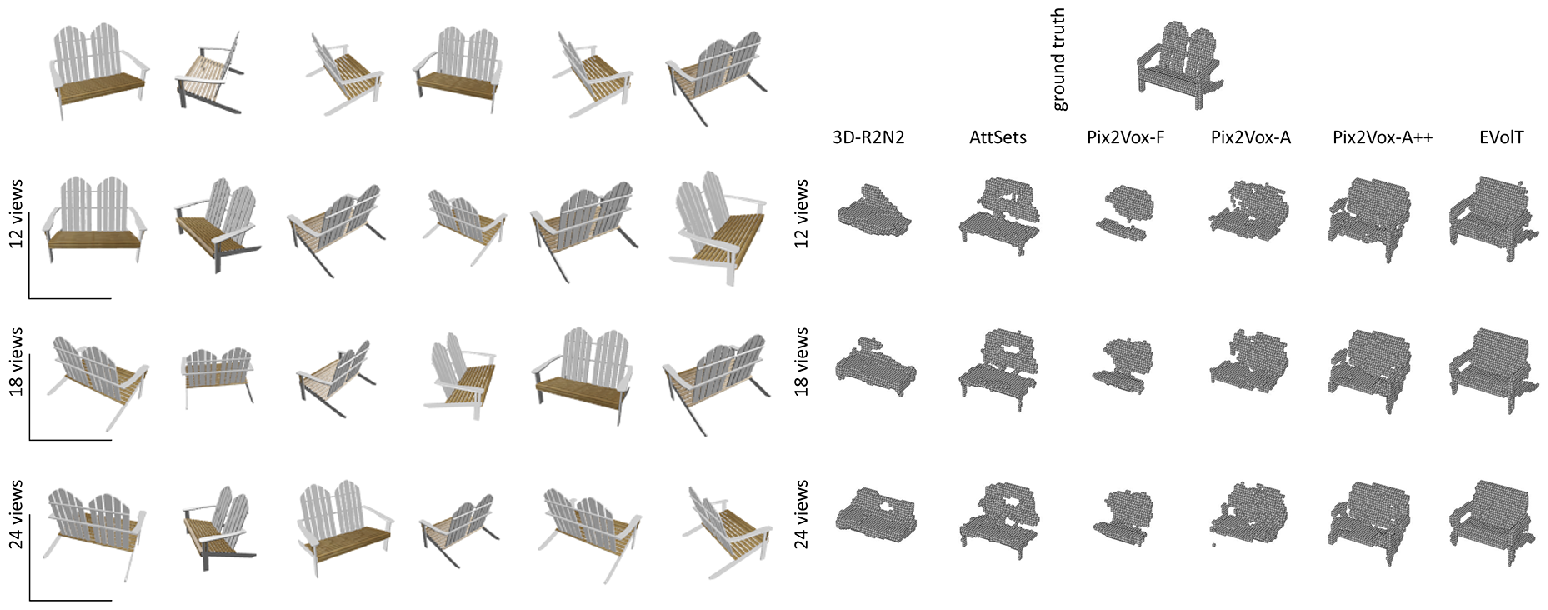} 
	\end{center}
	\caption{Qualitative reconstruction results of competing methods for sofa (top), aeroplane (middle), and bench (bottom).}
	\vspace{-0.05in}
	\label{fig:qual4}
\end{figure*}

We provide more object reconstruction results of competing methods, as shown in Figure \ref{fig:qual1}, \ref{fig:qual2}, \ref{fig:qual3}, and \ref{fig:qual4}.
In each object sample, we provide object reconstruction results from different number of input views, i.e., 12 views, 18 views, and 24 views.
The first two rows on the left part of Figure \ref{fig:qual1} show the 12 input views of an object, and the corresponding reconstruction results of competing methods are shown at the second row on the right.
Similarly, the first three rows on the left part are the 18 input views corresponding to the results on the right.
The qualitative comparison suggests the superiority of the proposed method in terms of the reconstruction topology and details.

\end{document}